\documentclass[11pt,a4paper]{article}
\usepackage{graphicx} % Required for inserting images
\usepackage{amsmath}
\usepackage{amsfonts}
\usepackage{dsfont}
\usepackage{algorithm}
\usepackage{algpseudocode}
\usepackage{amsmath}
\DeclareMathOperator*{\argmax}{arg\,max}

\RequirePackage{fancyhdr}

\RequirePackage{xcolor}

\RequirePackage{graphicx}

\newcommand{\ack}[1]{
\section*{Acknowledgments}}

\newcommand{\funding}[1]{
\section*{Funding}}

\newcommand{\data}[1]{
\section*{Data availability}}

\newcommand{\roles}[1]{
\section*{Author contributions}}

\newcommand{\suppdata}[1]{
\section*{Supplementary data}
}

\newcommand{\affil}[1]{{\fontsize{8}{10}\selectfont
       \raggedright }
	   }
\let\oldtabular\tabular
\renewcommand{\tabular}{\fontsize{8}{10}\selectfont \oldtabular}

\renewcommand{\title}[1]{{\exhyphenpenalty=10000\hyphenpenalty=10000 
 \fontsize{18}{21}\selectfont\noindent\raggedright
        \textsf{#1}\par}\suppressfloats[t]}

\renewcommand{\author}[1]{{\vspace{5mm}%
   \fontsize{10}{12}
      \raggedright #1
	  \vspace{3mm}}}

\newcommand{\email}[1]{\vspace*{12pt}{\fontsize{8}{10}\selectfont
       \raggedright {\bfseries E-mail:} #1 }
	  \vspace{3mm} }
	   
\newcommand{\keywords}[1]{{\fontsize{8}{10}\selectfont
       \raggedright {\bfseries Keywords:} #1}
	  }

\renewenvironment{abstract}{%
      \vspace{16pt plus3pt minus3pt}
      {\color{gray}\hrule} \ \\
      \noindent \fontsize{11}{12}\selectfont {\bfseries Abstract}\\
	  \rm\ignorespaces \raggedright}{\vspace{3mm} {\color{gray}\hrule}}

\usepackage{appendix}
\usepackage{subcaption}

\begin{document}

\title{Bayesian Sequential Optimal Experimental Design for Single-Pixel Imaging using Reinforcement Learning}
\author{Marcos Antonio Obando$^{1,2,*}$, Tianyuan Wang$^1$,\\ Tanja Tarvainen$^2$, Tristan van Leeuwen$^{1,3}$, Felix Lucka$^1$}
\date{\today}

% Options
\affil{$^1$Computational Imaging Group, Centrum Wiskunde \& Informatica (CWI), Amsterdam, The Netherlands}

\affil{$^2$Department of Technical Physics, University of Eastern Finland, Kuopio, Finland}

\affil{$^3$Matematisch Institute, Utrecht University, Utrecht, The Netherlands}

\affil{$^*$Author to whom any correspondence should be addressed.}

\email{Marcos.Obando@cwi.nl}


\begin{abstract}
    Adaptive compressed sensing (CS) aims to select the most informative measurements sequentially, which is crucial for many modern imaging systems. This task can be formulated as a Bayesian sequential optimal experimental design (sOED) problem, where each new acquisition is chosen to maximally reduce uncertainty about the unknown image given prior information and subsequent data provided by acquisitions. However, in single-pixel imaging (SPI), the large design space consists of all possible binary illumination patterns. Directly optimizing the Bayesian sOED objective over this space using conventional methods is generally challenging.
    
    In this work, we reformulate sOED for SPI as a reinforcement learning (RL) problem. In RL, an agent learns a policy through repeated experimental trials to maximize a reward, here given as the expected information gain, the most complete measure of uncertainty reduction in Bayesian OED. This avoids the need to directly optimize complex information-theoretic objectives during a measurement to obtain the next design. Instead, the result is a learned policy function that directly maps the current posterior belief about the image to the next optimal binary measurement pattern, in a computationally efficient way.

    Using a simple yet representative SPI imaging scenario, this work evaluates the scalability of the learned policy under two different policy parameterizations. Furthermore, we investigate how performance deteriorates when only approximate posterior information is available or the measurements are impaired by noise, and propose several strategies to introduce additional design constraints motivated by biomedical applications. The numerical experiments demonstrate that the proposed Bayesian sOED–RL framework remains robust and can be extended to more complex, large-scale SPI scenarios.
\end{abstract}

\section{Introduction}

Single-pixel imaging (SPI) emerged as an alternative to conventional array-based imaging systems, with a simplified hardware and operation in wavelengths where detector arrays are expensive or unavailable~\cite{duarte2008single} as their main advantages. Unlike traditional cameras, where an entire scene simultaneously using pixel arrays is captured, here an image is sequentially acquired as follows: A spatial light pattern illuminates the scene and a single photodetector records the total light intensity being reflected back. In practice, the patterns are structured via spatial light modulators or digital micromirror devices, the latter producing binary (on/off) illumination and making binary patterns a natural fit for SPI hardware~\cite{Zhang2017Hadamard,stojek2022single}. A clever choice of sensing patterns and reconstruction routines enable the recovery of images from far fewer measurements than the total number of pixels~\cite{donoho2006compressed,candes2006robust}. Known as \textit{compressed sensing} (CS), the mathematical framework display which properties should signals have so as to be exactly recovered from undersampled measurements, as long as the sensing patterns satisfy certain  conditions. 

Early approaches chose patterns for this CS framework in a generic manner, such as random binary projections or scrambled Hadamard matrices~\cite{gan2008fast}, which provide reasonable results over a broad range of images. When prior knowledge about the samples that are images is available, one can improve the results by adapting the acquisitions. In an \textit{optimal experimental design} (OED) framework, a representation of the possible images enables the selection of maximally informative measurement patterns, allowing an efficient image recovery~\cite{rainforth2024modern}. 

While conventional OED determines a fixed set of optimal patterns before any measurements are taken, \textit{sequential} OED (sOED) proposes a way more efficient strategy~\cite{shen2023bayesian}. Instead of deploying all patterns upfront, sOED adaptively selects each new pattern \textit{during} the measurement process, based on what has already been learned from previous measurements. As early measurements mainly localize where interesting features lie, later ones focus on details around those specific regions. The 'a-posteriori' distribution, a Bayesian update of the prior knowledge by each data acquisition, results in a natural state representation that captures all available information.

However, sOED for imaging faces a severe computational challenge. On each step, one must solve an optimization problem over all possible binary measurement patterns to find the one that maximally reduces the uncertainty about the image. For an image with $M$ pixels, the $2^M$ possible binary patterns make conventional per-step optimization computationally prohibitive, even at modest image sizes~\cite{welch1982branch,foster2021deep}, limiting the practical applicability of sOED to real-world imaging problems. 

To overcome this barrier, we pose sOED as a \textit{reinforcement learning} (RL) problem. In this approach, rather than solving an optimization problem on each step, one looks for a parametric policy mapping the current posterior distribution directly to measurement patterns. Using the expected information gain between prior and posterior image knowledge as the reward signal, policy training consists on selecting the most informative patterns given what has already been observed along experiments. At evaluation time, the policy acts as a pattern proposal function, without needing any further optimization. 

In this work, two kinds of policy parameterizations are proposed and their scalability is studied. A \textit{categorical policy} assigns a probability to each of the $2^M$ possible binary patterns, whereas a \textit{element-wise} Bernoulli policy sets each pixel's activation probability independently. Under a controlled SPI scenario, where the optimal policy is known and information gain can be computed exactly, their performance is compared after training against random and optimal design strategies. Moreover, the replacement of the posterior distribution as a state variable by a sample-based approximate and the addition of noise to the measurements are analyzed, as well as hard and soft energy constraints over the binary projections are proposed.

The paper is organized as follows. Section \ref{sec:Background} presents the elements of single-pixel imaging and its treatment as an inverse problem, how Bayesian optimal experimental design can be applied to it and its close relation to sequential decision making processes. Section \ref{sec:Methods} bridges this theory using a reinforcement learning abstraction, and illustrates the technical implementation details for the single-pixel imaging case. Section \ref{sec:NumericalStudies} introduces setting considerations for the proposed learning scenario and numerical results obtained in the studies, with a discussion in Section \ref{sec:Discussion} and conclusions in Section \ref{sec:Conclusions}.

\section{Background} \label{sec:Background}
\subsection{Single-pixel imaging as a Bayesian inverse problem}\label{sec:SPIasBIP}

A single measurement in SPI can be described as 
\begin{equation}\label{eq:SingleMeasurement}
    y =  \langle \mathbf{h}, \mathbf{x} \rangle + \varepsilon, 
\end{equation}
where $\mathbf{x}\in \mathbb{R}^{M}$ is the image that one wants to recover in vector form, $\mathbf{h}_N \in \{0, 1\}^{M}$ is a discrete binary pattern and $\varepsilon$ is a noise realization, usually modeled as a normally distributed uncorrelated random variable $\mathcal{N}(0, \sigma^2)$ with mean zero and standard deviation $\sigma$. The SPI forward model, collecting $N$ observations $\mathbf{y}_N\in\mathbb{R}^N$, is described in a matrix-vector form as
\begin{align}\label{eq:SPIInverse}
    \begin{split}
    \mathbf{y}_N &= \mathbf{H}_N\mathbf{x} + \boldsymbol{\varepsilon}_N\\
        \begin{bmatrix}
    y_1  \\
    \vdots\\
    y_N
\end{bmatrix}
 &= 
\begin{bmatrix}
    \mathbf{h}_1^T  \\ \vdots\\ \mathbf{h}_N^T
\end{bmatrix}
\begin{bmatrix}
    x_1  \\
    \vdots\\
    x_M
\end{bmatrix}
+
\begin{bmatrix}
    \varepsilon_1  \\
    \vdots\\
    \varepsilon_N
\end{bmatrix}
    \end{split} \quad ,
\end{align}
with $\mathbf{H}_N$ being the forward operator, which rows consist of discrete binary patterns $\mathbf{h}_n$. Assuming that the noise from individual measurements is uncorrelated, the noise variable $\boldsymbol{\varepsilon}_N$ follows a multivariate normal distribution $\boldsymbol{\varepsilon}_N\sim\mathcal{N}(\mathbf{0}_{N}, \sigma^2 \mathbf{I}_N)$. When $N < M$, the inverse problem of recovering $\mathbf{x}$ from measurements $\mathbf{y}_N$ is underdetermined, and usually ill-conditioned, which means that a perturbation in the image $\mathbf{x}$ does not necessarily lead to an identifiable change in the observed data $\mathbf{y}_N$~\cite{scherzer2009variational}. 

To prevent this, assumptions based on 'a-priori' knowledge of the image can be used to regularize the possible solutions of the problem. In combination with how consistent measurements are with respect to the forward model, an updated, 'a-posteriori' knowledge of the image is obtained. This methodology can be formalized using the Bayesian framework,  where $\mathbf{x}$ is treated as a random variable rather than merely a real-valued vector, describing the solution to Eq. \eqref{eq:SPIInverse} through the Bayes formula~\cite{stuart2010inverse}

\begin{equation}\label{eq:Bayes}
    P(\mathbf{x}\mid\mathbf{y}_N; \mathbf{H}_N) = \frac{P(\mathbf{y}_N\mid\mathbf{x}; \mathbf{H}_N)P(\mathbf{x})}{P(\mathbf{y}_N; \mathbf{H}_N)}
\end{equation}

% : \mathcal{X} \rightarrow [0,1]\in\mathbb{P}_\mathbf{x}
Here, the prior distribution $P(\mathbf{x})$ encompasses the knowledge of the image before  measurement takes place and can encode the properties one expects the image to have, e.g., total variation priors were proposed to model images with clearly defined edges between piecewise constant areas \cite{lassas2004can}.  Note that the prior distribution is independent of the measurement operator $\mathbf{H}_N$. The likelihood distribution $P(\mathbf{y}_N\mid\mathbf{x}; \mathbf{H}_N)$ quantifies the probability of measuring $\mathbf{y}_N$, which contains all measurements up to step $N$, given the image $\mathbf{x}$ using the measurement operator $\mathbf{H}_N$. Here the noise and the unknown $\mathbf{x}$ are mutually independent variables

\begin{align}\label{eq:Likelihood}
    \begin{split}
     P(\mathbf{y}_N\mid\mathbf{x}; \mathbf{H}_N) &\propto \exp \left( - \tfrac{1}{2 \sigma^2} \| \mathbf{H}_N \mathbf{x} - \mathbf{y}_N  \|_2^2 \right) \\ &= \prod_{n=1}^N  \exp \left( - \frac{(\mathbf{h}^T_n  \mathbf{x} - y_n)^2}{2 \sigma^2}\right) =: \prod^{N}_{n=1} P(y_n|\mathbf{x}; \mathbf{h}_n)
    \end{split}
\end{align}
The marginal likelihood $P(\mathbf{y}_N; \mathbf{H}_N)$, also known as model evidence, is a normalizing factor for all possible images $\mathbf{x}$ given a particular $\mathbf{H}_N$
\begin{equation}\label{eq:ModelEvidence}
P(\mathbf{y}_N; \mathbf{H}_N) = \mathbb{E}_{P(\mathbf{x})}\big[P(\mathbf{y}_N|\mathbf{x}; \mathbf{H}_N)\big]   
\end{equation}
As a result, the posterior distribution, $P(\mathbf{x}\mid \mathbf{y}_N; \mathbf{H}_N)$, combines information from the prior and likelihood distributions, assigning a probability score to each possible image $\mathbf{x}$ given observed data $\mathbf{y}_N$. Note that due to the sequential, independent SPI measurement process, the posterior distribution is updated during acquisition as
\begin{align}\label{eq:BayesSequential}
    \begin{split}
        P(\mathbf{x}|\mathbf{y}_N; \mathbf{H}_N)& \propto P(\mathbf{y}_N|\mathbf{x}; \mathbf{H}_N)P(\mathbf{x})\\
        & \overset{\substack{\eqref{eq:Likelihood}}}{\propto}  P(y_{N}|\mathbf{x};\mathbf{h}_N)P(\mathbf{y}_{N-1}|\mathbf{x}; \mathbf{H}_{N-1})P(\mathbf{x})\\
        & \propto P(y_{N}|\mathbf{x}; \mathbf{h}_N)P(\mathbf{x}|\mathbf{y}_{N-1}; {\mathbf{H}_{N-1}})
    \end{split}
\end{align}

\subsection{Bayesian optimal experimental design}\label{sec:BED}

The aim in compressed SPI is to obtain a useful reconstruction of $\mathbf{x}$ using $N < M$ measurements. An important question is which of the $2^{NM}$ possible \emph{experimental designs} $\mathbf{H}_N \in \{0,1\}^{N \times M}$ one should choose for a given prior $P(\mathbf{x})$. Bayesian OED formalizes this problem: First, it uses the Shannon entropy as an information-theoretic measure to quantify the uncertainty about the random image $\mathbf{x}$ before the measurement 
\begin{equation}\label{eq:Entropy}
    Ent(\mathbf{x}) := -\mathbb{E}_{P(\mathbf{x})}\Big[\log P(\mathbf{x})\Big]
\end{equation}
Then, it assesses how informative a particular collection of measurements $\mathbf{y}_N$ acquired with a design $\mathbf{H}_{N}$ is by introducing the \textit{information gain} $\mathrm{IG}_\mathbf{x}(\mathbf{H}_{N}, \mathbf{y}_{N})$ as a difference between prior and posterior entropy

\begin{align}\label{eq:IG}
        \mathrm{IG}_\mathbf{x}(\mathbf{y}_{N}; \mathbf{H}_{N}) := Ent(\mathbf{x}) -Ent(\mathbf{x}\mid\mathbf{y}_{N}; \mathbf{H}_{N})
\end{align}

Using the marginal likelihood in \eqref{eq:ModelEvidence} and the information gain in \eqref{eq:IG}, the \textit{expected information gain} $\mathrm{EIG}_{\mathbf{x}}(\mathbf{H})$ provides an average uncertainty change with respect to all possible measurement realizations $\mathbf{y}_N$
% Information gain utility computation produced by an arbitrary design presents a challenging double integration: a) an outer expectation averaging over the $\mathcal{X}$ image set and b) an inner expectation quantifying the likelihood of possible outcomes from the over the measurement set $\mathcal{Y}$.

\begin{align}\label{eq:EIG}
        \mathrm{EIG}_\mathbf{x}(\mathbf{H}_{N}) &:= \mathbb{E}_{P(\mathbf{y}_{N};\mathbf{H}_{N})}\big[\mathrm{IG}_\mathbf{x}(\mathbf{y}_{N};\mathbf{H}_{N})]
\end{align}

BOED chooses the measurement patterns that maximize the expected information gain

\begin{equation}\label{eq:PrD}
    \mathbf{H}_N^{\rm Pr} := \argmax_{\mathbf{H}_N  \in \{0,1\}^{N \times M}} \mathrm{EIG}_{\mathbf{x}}(\mathbf{H}_{N}),
\end{equation}

where $\mathbf{H}_N^{\rm Pr}$ refers to an '\textit{a-priori}' design , as this choice only relies on information available before the measurement, i.e., the prior. Now consider the case where one can design pattern sequentially during the measurement, based on the information gathered so far, i.e., '\textit{a-posteriori}' (sequential Optimal Experimental Design, sOED). In Bayesian sOED, consider the expected information gain of using pattern $\mathbf{h}_n$ for the $n^{th}$ measurement step given that the previous $\mathbf{y}_{n-1}$ measurements were acquired with the design $\mathbf{H}_{n-1}$

\begin{align}\label{eq:SequentialIG}
    \mathrm{EIG}_{\mathbf{x}\mid \mathbf{y}_{n-1};\mathbf{H}_{n-1}}(\mathbf{h}_{n}) := \mathbb{E}_{P(y_n;\mathbf{h}_n)}\big[\mathrm{IG}_\mathbf{x\mid\mathbf{y}_{n-1};\mathbf{H}_{n-1}}(y_{n};\mathbf{h}_{n})\big], \quad 
\end{align}

where the marginal likelihood can be expressed as an expectation over the new measurement $y_n$ likelihood using the the posterior until step $n-1$

\begin{align}\label{eq:OneStepModelEvidence} 
    P(y_n ; \mathbf{h}_n) := \mathbb{E}_{P(\mathbf{x}\mid \mathbf{y}_{n-1}; \mathbf{H}_{n-1})}[P(y_n\mid\mathbf{x};  \mathbf{h}_n)], \quad P(\mathbf{x}\mid\mathbf{y}_0; \mathbf{H}_0) := P(\mathbf{x}) 
\end{align}

Note that the prior $P(\mathbf{x})$ in \eqref{eq:IG}, \eqref{eq:ModelEvidence} was merely exchanged for the posterior $P(\mathbf{x}\mid \mathbf{y}_{n-1}; \mathbf{H}_{n-1})$ in \eqref{eq:SequentialIG}, \eqref{eq:OneStepModelEvidence} and only consider the design of the following pattern $\mathbf{h}_n$. The \emph{greedy} sOED $\mathbf{H}^{\rm Gr}_N(\mathbf{x})$, which compared to \eqref{eq:PrD} depends on the unknown realization $\mathbf{x}$, maximizes \eqref{eq:SequentialIG} in every step

\begin{equation}\label{eq:GreedySOED}
    \mathbf{h}_n^{\rm Gr}(\mathbf{x}) := \argmax_{\mathbf{h}_n  \in \{0,1\}^{M}} \mathrm{EIG}_{\mathbf{x | \mathrm{y}_{n-1}; \mathrm{H}_{n-1}}}(\mathbf{h}_n) , \quad \mathbf{H}^{\rm Gr}_N = [\mathbf{h}^{\rm Gr}_1, \ldots, \mathbf{h}^{\rm Gr}_N]^T
\end{equation}

Greedy designs reduce the chain of sequential design decisions that all influence the informativeness of the final $N$-step design $\mathbf{H}_N$ into a sequence of isolated design decision steps. Although this is often computationally attractive, it is a suboptimal solution to the problem of maximizing the sum of expected information gains over all $N$ steps \cite{nemhauser1978analysis} \cite{alexanderian2026submodularity}. As an alternative, the \textit{full sequential OED} $\mathbf{H}_N^{\rm F}(\mathbf{x})$ is the globally optimal sequence of measurement patterns $\mathbf{h}_1, \ldots, \mathbf{h}_N$ that maximizes the cumulative expected information gain across all $N$ steps

\begin{align}\label{eq:sOED}
    \mathbf{H}_N^{\rm F}(\mathbf{x}) := \argmax_{\mathbf{h}_1, \ldots, \mathbf{h}_N \in \{0,1\}^{M}} \sum_{n=1}^{N} \mathrm{EIG}_{\mathbf{x}\mid \mathbf{y}_{n-1}; \mathbf{H}_{n-1}}(\mathbf{h}_{n})
\end{align}

Unlike greedy approaches that optimize each measurement independently, this formulation accounts for the long-term information-gathering strategy by jointly optimizing over the entire measurement sequence. The key insight is that early measurement choices influence the posterior distributions encountered in subsequent steps, creating a complex interdependence between decisions that greedy methods fail to capture. 

The theoretical superiority of the full sOED strategy with respect to the greedy approach, introduced in \cite{shen2023bayesian}, is formalized as

\begin{align}\label{eq:OEDcmp}
    \sum_{n=1}^{N} \mathrm{EIG}_{\mathbf{x}\mid \mathbf{y}_{n-1}; \mathbf{H}^{\rm F}_{n-1}}(\mathbf{h}^{\rm F}_{n}) &\geq \sum_{n=1}^{N} \mathrm{EIG}_{\mathbf{x}\mid \mathbf{y}_{n-1}; \mathbf{H}^{\rm Gr}_{n-1}}(\mathbf{h}^{\rm Gr}_{n})\\
    \sum_{n=1}^{N} \mathrm{EIG}_{\mathbf{x}\mid \mathbf{y}_{n-1}; \mathbf{H}^{\rm F}_{n-1}}(\mathbf{h}^{\rm F}_{n}) &\geq \mathrm{EIG}_{\mathbf{x}}(\mathbf{H}^{\rm Pr}_{N})\label{eq:OEDpri}
\end{align}

Inequality \eqref{eq:OEDcmp} states that the cumulative information gain achieved by the full sOED sequence $\mathbf{H}_N^{\rm F}$ is at least as large as that obtained by the greedy approach $\mathbf{H}_N^{\rm Gr}$, which myopically maximizes the EIG at each individual step without considering future measurements. This gap arises because the full sOED can strategically select measurements that may have lower immediate information gain but enable more informative measurements in subsequent steps a form of temporal credit assignment that greedy methods cannot perform. Inequality \eqref{eq:OEDpri} further shows that the full sOED outperforms the a-priori design $\mathbf{H}_N^{\rm Pr}$, which selects all $N$ patterns before observing any data and thus cannot adapt to the information revealed by early measurements. Together, these inequalities establish the full sOED as the optimal approach within the Bayesian framework, motivating the development of tractable approximation methods such as the RL-based approach presented in this paper.

\textbf{Computational considerations.} While the full sOED \eqref{eq:sOED} is theoretically superior to the a-priori OED design \eqref{eq:PrD}, it comes with similar computational challenges: Taking the interdependence of the terms of the sum in \eqref{eq:sOED} into account means that one also needs to optimize over the exponentially large space $\{0,1\}^{NM}$ of binary patterns. Even for modest image sizes (e.g., $M=64$), the design space contains $2^{64} \approx 10^{19}$ possible patterns at each step, making exhaustive search or traditional optimization methods computationally prohibitive.

Moreover, computing the expected information gain $\text{EIG}_{\mathbf{x}|\mathbf{y}_{n-1}; \mathbf{H}_{n-1}}(\mathbf{h}_n)$ in \eqref{eq:SequentialIG} requires integrating over all possible measurement outcomes through the marginal likelihood $P(y_n ; \mathbf{h}_n)$ in \eqref{eq:OneStepModelEvidence}, which itself involves an expectation over the posterior distribution from previous measurements. In practice, these integrations are often high-dimensional and lack closed-form expressions, relying on Monte Carlo-based approximations (see Section \ref{sec:RewardApprox}). This computation is repeated at each of the $N$ sequential steps, making the full sOED approach \eqref{eq:sOED} challenging for real-time or resource-constrained imaging applications.

These computational challenges motivate the reformulation of the sOED problem as a sequential decision-making under uncertainty, and the introduction of RL-based approaches as a principled way to tackle the sOED problem \eqref{eq:sOED}. Instead of solving the complex optimization problems behind sOED, one learns parametric policies that directly map posterior distributions to near-optimal measurement patterns by simulating a large number of sOED instances offline, i.e. before any real measurement takes place. The complex integrals only have to be computed during this offline training phase, not during deployment.

% &= \argmax_{\mathbf{H}_N}\int_\mathbf{\mathcal{X}}\int_\mathbf{\mathcal{Y}} \mathrm{I}_{\mathbf{x}}(\mathbf{H}_N,\mathbf{y}_N)P(\mathbf{y}_N\mid \mathbf{x}, \mathbf{H}_N)P(\mathbf{x})d\mathbf{y}_Nd\mathbf{x}\\

% Expected information gain serves as a proper utility function, evaluating how informative an acquisition operator is with respect to the image under study \cite{lindley1956measure, bernardo1979expected}. For single-pixel imaging BOED, an optimal design $\mathbf{H}_N^{*}$ is the one maximizing the prior to posterior expected information gain, as defined in \eqref{eq:IG_Sequential}:

% \begin{align}\label{eq:UtilityOED}
%     \begin{split}
%     \mathbf{H}_N^* &= \argmax_{\mathbf{H}_N} \mathrm{EIG}_{\mathbf{x}}(\mathbf{H}_N)\\
%      &= \argmax_{[\mathbf{h}_1, \ldots, \mathbf{h}_N]} \sum_{n=1}^{N} \mathrm{EIG}_{\mathbf{x}\mid \mathbf{y}_{n}, \mathbf{H}_{n-1}}(\mathbf{h}_{n})\\
%     % \mathbf{H}_N^* &= \argmax_{\mathbf{H}_N = [\mathbf{h}_1, \ldots, \mathbf{h}_N]} \sum_{n=1}^{N} I_{\mathbf{x}\mid \mathbf{y}_{n}, \mathbf{H}_{n-1}}(\mathbf{h}_{n}) \\
%     \end{split}
% \end{align}

\subsubsection{Sequential decision-making processes}\label{sec:POMDP}

% Describe sOED as a generic sequential decision process under uncertainty, and introduce the model of a Markovian Markov decision process and an POMDP
Solving the sOED problem \eqref{eq:sOED} can be seen as an instance of a \textit{Markov decision process} (\textit{MDP}), a mathematical model to describe sequential and time-discrete decision processes under uncertainty. In MDPs, an \textit{agent} interacts dynamically with an \textit{environment} in order to accumulate a \textit{reward}: The \textit{state} $S \in \mathcal{S}$ represents the current configuration of the agent and environment, and every possible decision the agent can make is represented by an \textit{action} $A \in \mathcal{A}$ that will lead to a new state $S'$ with a certain probability. The agent's decision making is represented by a \textit{policy}  $\pi(S,A)$, which is a probability distribution in $A$, encoding the likelihood that the agent takes the decision $A$ when in state $S$. For every decision, the agent receives a \textit{reward} $ R(S, A, S')$. The goal in MDP is to find policies that maximize the accumulated rewards for a given time window. Very often, the agent does not have full knowledge of the state $S$, leading to a \textit{partially observable Markov decision process} (\textit{POMDP}) \cite{SuBa2018}.

%In a sequential decision-making process, interactions between an agent and an environment are linked with a temporal ordering. The configuration of the environment is encoded with \textit{state} $S$ and the possible decisions the agent can make are formalized as \textit{actions} $A$. For a Markovian process, an agent observes the current state of the environment and selects an action according to a certain \textit{policy} $\pi$, namely a numerical frame of its decision strategy. Due to this, the environment transitions to a new state $S'$ and the agent receives a numerical feedback, known as \textit{reward} $R$, which indicates the performance of the agent. The reward relates to the task assigned to the agent and can be used to obtain better decision-making policies.

% Introduce Reinforcement Learning as a framework to find optimal policies for POMDP and the exploration–exploitation dilemma. 

Several methodologies describe how to find a policy that maximizes the sum of rewards from the starting time $n =1$ until the end of the experiment. In complex scenarios, obtaining analytical solutions for this objective or using conventional optimization schemes is often infeasible. Instead, RL methods achieve the optimization goal by updating the agent's policy based on the feedback provided by the environment at each time. The learning process accounts not only for \textit{exploiting} stepwise successful actions, but also by \textit{exploring} the landscape of possible environment states~\cite{SuBa2018}. Two principal classes of RL algorithms exist: \textit{value-based methods} and \textit{direct policy search} methods. In \textit{value-based methods}, the agent learns a scalar estimate of the expected cumulative rewards from a given state onward under a fixed policy, called the \textit{value function}, from repeated environment interactions and derives a policy by acting greedily with respect to this function. In \textit{direct policy search}, the policy is explicitly represented by a parametric function $\pi_\theta$ and $\theta$ is updated directly via stochastic gradient ascent, i.e.\ \textit{policy gradient methods}~\cite{williams1992simple,sutton2000policy}.

 In the sOED setting, however, the practical benefits of RL are only realized when the agent can be trained on a sufficiently large number of simulated trajectories, so that it can repeatedly experience the stochastic transitions $S \to S'$ and the associated rewards $R(S,A,S')$ under many different policies. If the available simulation budget is too small, the agent cannot adequately explore the joint state-action space, and the resulting policy is prone to overfitting to a few sampled scenarios rather than approximating the truly optimal policy of the underlying MDP/POMDP, which substantially limits the applicability of RL-based sOED in practice.

\section{Methods} \label{sec:Methods}

\subsection{SPI-sOED reinforcement learning framework}\label{sec:RL-SPI-sOED}

In this section, the sequential optimal experimental design problem for single-pixel imaging defined in Eq.~\eqref{eq:sOED} is formulated as a RL task and a methodology to solve it is developed. The key insight is that one must choose a complete sequence of patterns that jointly maximizes the cumulative information gain about the unknown image over the entire experiment, accounting for how each measurement shapes future decisions.

 The Bayesian framework from Sec.~\ref{sec:SPIasBIP} induces this sequential structure: At any measurement step $n$, the current knowledge about the image $\mathbf{x}$ is summarized by the posterior distribution $P(\mathbf{x}\mid\mathbf{y}_{n-1}; \mathbf{H}_{n-1})$, where $\mathbf{y}_{n-1}$ contains all measurements up to step $n-1$ and $\mathbf{H}_{n-1}$ contains all previously used patterns. This reduces to the prior $P(\mathbf{x})$ when $n=0$ (see Equation ~\eqref{eq:ModelEvidence}). When a new pattern $\mathbf{h}_n$ is chosen and measurement $y_n$ is obtained, Bayes' rule gives us an updated posterior $P(\mathbf{x}\mid\mathbf{y}_n; \mathbf{H}_n)$, which represents the refined knowledge. This sequential updating process can be outlined using the following decision-theoretic components:

\begin{itemize}
\item \textbf{State:} The posterior distribution $P(\mathbf{x}\mid\mathbf{y}_{n-1}; \mathbf{H}_{n-1})$ after measurement $n-1$ serves as the current state $S_n$. This state encodes all information from past measurements and patterns that is relevant for future decisions.

\item \textbf{Action:} The binary measurement pattern $\mathbf{h}_n$ used at step $n$ becomes the action $A_n$. It leads to the scalar measurement $y_n = \mathbf{h}_n^\top \mathbf{x} + \epsilon_n$.

\item \textbf{State transition:} The new posterior distribution $P(\mathbf{x}\mid\mathbf{y}_n; \mathbf{H}_n)$ after measurement $n$ becomes the next state $S_{n+1}$. This transition is stochastic because the measurement $y_n$ is corrupted by noise $\epsilon_n$, but the Bayesian update rule provides the deterministic mapping from $(S_{n}, A_{n}, y_n)$ to $S_{n+1}$.

\item \textbf{Reward:} The immediate reward for choosing pattern $\mathbf{h}_n$ in state $S_n$ is the expected information gain $R_n := \mathrm{EIG}_{\mathbf{x}\mid\mathbf{y}_{n-1}; \mathbf{H}_{n-1}}(\mathbf{h}_n)$, defined in Eq.~\eqref{eq:EIG}. Note that this reward depends only on the current state $S_n$ and action $A_n$, and not on the resulting new state $S_{n+1}$. Computing $\mathrm{EIG}$ requires marginalizing over all possible outcomes $y_n$, which is only tractable in small-scale settings as the one showcased on this work. For more complex settings, $R_n$ can be approximated by the actual information gain $\mathrm{IG}_{\mathbf{x}\mid\mathbf{y}_{n-1};\mathbf{H}_{n-1}}(\mathbf{h}_n, y_n)$, which depends on the observed $y_n$ and thereby $S_{n+1}$, cf. Section \ref{sec:RewardApprox}.

\item \textbf{Episode:} A complete sequence of $N$ measurement steps forms an episode or trajectory $\tau := \big(S_0, A_1, S_1, R_1, \ldots, S_{N-1}, A_N, S_N, R_N\big)$, representing one complete imaging experiment from prior to final posterior.

\item \textbf{Policy:} The stochastic policy $\pi(A_n \mid S_n)$ maps the current state, the posterior distribution over $\mathbf{x}$, to a probability distribution over binary measurement patterns. 
\end{itemize}

With these definitions, the sequential optimal design problem of maximizing the cumulative expected information gain from Eq.~\eqref{eq:sOED} can be recast as finding an optimal policy $\pi^*$ that maximizes the total reward
\begin{align}\label{eq:RL-sOED}
\begin{split}
\pi^* :&= \argmax_{\pi} \mathbb{E}_{\tau \sim P(\tau \mid \pi)} \Bigg[\sum_{n=1}^{N} R_n \Bigg] \\&= \argmax_{\pi} \mathbb{E}_{\tau \sim P(\tau \mid \pi)} \Bigg[\sum_{n=1}^{N} \mathrm{EIG}_{\mathbf{x}\mid \mathbf{y}_{n-1}; \mathbf{H}_{n-1}}(\mathbf{h}_n)\Bigg] \\&=: \argmax_{\pi} J(\pi)
\end{split}
\end{align}
where $J(\pi)$ is defined as the objective function for the policy training. The expectation in \eqref{eq:RL-sOED} over the distribution of trajectories $\tau \sim P(\tau \mid \pi)$ integrates the stochasticity of the sample, policy and the observation noise. A trajectory can be generated as follows: a sample is drawn from the prior $\mathbf{x} \sim P(\mathbf{x})$, set $S_0 =P(x)$ and iterate for $n=1, \ldots, N$

\begin{equation}\label{eq:trajectory}
\begin{aligned}
&\quad A_n = \mathbf{h}_n \sim \pi(\cdot\mid S_{n-1}), \\
&\quad y_n \sim P(y_n\mid\mathbf{x};\, \mathbf{h}_n), \\
&\quad S_n = P(\mathbf{x}\mid\mathbf{y}_n;\, \mathbf{H}_{n-1}) \propto P(y_n\mid\mathbf{x};\,\mathbf{h}_n)\, P(\mathbf{x}\mid\mathbf{y}_{n-1};\,\mathbf{H}_{n-1}), \\
&\quad R_n = \mathrm{EIG}_{\mathbf{x}\mid\mathbf{y}_{n-1};\,\mathbf{H}_{n-1}}(\mathbf{h}_n) = \mathbb{E}_{P(y_n;\,\mathbf{h}_n)}\!\big[\mathrm{IG}_{\mathbf{x}\mid\mathbf{y}_{n-1};\,\mathbf{H}_{n-1}}(y_n;\,\mathbf{h}_n)\big].
\end{aligned}
\end{equation}

The remaining question is how to represent and optimize the policy $\pi$ in practice. The sOED state space consists of full posterior distributions over $\mathbf{x}$, which are high-dimensional and continuous, making value function estimation intractable. This motivates the use of \textit{direct policy search} methods, where $\pi$ is parameterized by a differentiable function $\pi_\theta$ and optimized end-to-end via policy gradient ascent on $J(\theta)$. In the following sections, the specific parameterizations of $\pi$ adopted for SPI-sOED and the resulting learning algorithm are detailed.

\subsection{SPI design policy parameterizations}\label{sec:PolicyArchitectureComparison}

The fundamental question is how to represent this distribution over binary patterns. To construct a policy from the current posterior state, we use a feed-forward neural network that maps a state representation $S_{n-1}$ to a probability distribution over binary measurement patterns. We consider two policy parameterizations that share the same multilayer perceptron (MLP) backbone but differ in their output dimension and final activation function. The MLP follow the model

\begin{equation}
f_\theta := w_{\theta_{L-1}} \circ \sigma_{L-1} \circ \ldots \circ \sigma_1 \circ w_{\theta_1},
\end{equation}

where $w$ denotes an affine linear mapping with parameters $\theta_l$, $\sigma_l(\cdot)$ a non-linear activation function and $\circ$ the composition of functions such that $(f\circ g)(\cdot) = f(g(\cdot))$, with $l=1,\ldots,L-1$ the layer index. 

A categorical policy $\pi^C_\theta$ directly outputs a probability distribution over all $2^M$ possible patterns. With an output layer of $2^M$ units and a Softmax activation, the probability for a possible pattern $\mathbf{h}_i$ given a particular past state $S_{n-1}$ is

\begin{equation}\label{eq:CategoricalPolicy}
\pi^C_\theta( \mathbf{h}_i | S_{n-1}) = \frac{e^{f_\theta(S_{n-1})_i}}{\sum_j e^{f_\theta(S_{n-1})_j}} ,
\end{equation}
where $j=1,\ldots,2^M$ indexes all possible patterns. This approach allows the network to learn arbitrary correlations between pixel activations. However, it scales exponentially in $M$, making it practical only for very small images.

Alternatively, each pixel activation can be modeled independently: with $M$ output units and a sigmoid activation $\sigma_L$, we obtain independent Bernoulli probabilities for the activation of each pixel $h_k, k=1,\ldots, M$ of a pattern $\mathbf{h}$

\begin{equation}\label{eq:BernoulliPolicy}
\pi^B_\theta(h_k = 1| S_{n-1}) = \text{Bernoulli} \left( \frac{1}{1 + e^{-f_\theta(S_{n-1})_k}} \right).
\end{equation}

To sample a pattern, independent draws are sampled for each pixel from its corresponding Bernoulli distribution. This scales linearly in $M$, making it practical for larger images. Although the network still can learn to modulate individual probabilities based on the full posterior state, it cannot model correlations between pixel activations directly. With these state representations in place, the architecture of the input layer is determined by how the state is encoded, since the model processes that representation in a compatible form as input.

\subsection{State representation and approximation}\label{sec:StateApprox}

The RL framework from Sec.~\ref{sec:RL-SPI-sOED} requires at each step $n$ a state $S_n$ merging the current knowledge about the image and a reward $R_n$ quantifying information gained. In the SPI-sOED setting, both quantities are obtained via the Bayesian posterior

\begin{equation}
    P(\mathbf{x} \mid \mathbf{y}_{n-1}; \mathbf{H}_{n-1}),
\end{equation}
but their computation depends on whether this posterior is available in closed-form or only implicitly through sampling. As we introduced in Section \ref{sec:POMDP}, using a stochastic variable as a state corresponds to a partially observable Markov decision process.

\paragraph{Explicit posterior (closed-form state).}
When the posterior admits a closed-form expression, the full distribution reads
\begin{equation}
    S_n \;=\; P(\mathbf{x} \mid \mathbf{y}_{n-1}; \mathbf{H}_{n-1})
    \label{eq:StateExplicit}
\end{equation}
as the state.

A typical example is a linear Gaussian model with prior $\mathbf{x} \sim \mathcal{N}(\boldsymbol{\mu}_0, \mathbf{\Sigma}_0)$ and likelihood

\begin{equation}
    y_n \mid \mathbf{x}, \mathbf{h}_n \sim \mathcal{N}(\mathbf{h}_n^\top \mathbf{x}, \sigma^2),
\end{equation}

for which the posterior is again Gaussian,

\begin{align}
    \mathbf{\Sigma}_n 
    &= \Bigl(\mathbf{\Sigma}_{n-1}^{-1} + \sigma^{-2}\,\mathbf{h}_n \mathbf{h}_n^\top \Bigr)^{-1}, \label{eq:GaussianPosteriorCov}\\
    \boldsymbol{\mu}_n 
    &= \mathbf{\Sigma}_n\Bigl(\mathbf{\Sigma}_{n-1}^{-1}\boldsymbol{\mu}_{n-1} + \sigma^{-2}\,y_n\,\mathbf{h}_n \Bigr),
    \label{eq:GaussianPosteriorMean}
\end{align}
with $(\boldsymbol{\mu}_n, \mathbf{\Sigma}_n)$ fully characterising $S_n$. In this regime, both the state and the expected information gain can be evaluated exactly, providing a high-fidelity learning signal.

\paragraph{Implicit posterior (sample-based state).}
For more complex priors or likelihoods, the posterior $P(\mathbf{x} \mid \mathbf{y}_{n-1}, \mathbf{H}_{n-1})$ is not available in closed-form and must instead be represented implicitly via a collection of posterior samples $\{\mathbf{x}^{(i)}\}_{i=1}^{N_s} \sim P(\mathbf{x} \mid \mathbf{y}_{n-1}, \mathbf{H}_{n-1})$. Among several possible ways to encode this implicit representation as a state, two natural choices are considered. The first is to pass a set of samples directly to the policy,

\begin{equation}
    \hat{S}_n \;=\; \bigl\{\mathbf{x}^{(i)}\bigr\}_{i=1}^{N_s},
    \label{eq:StateSamples}
\end{equation}

which preserves the complete distributional information at the cost of a fixed-size input of dimension $N_s \times M$. The second, computationally cheaper alternative is to summarize the set by its empirical mean and covariance analogues of the sufficient statistics $(\boldsymbol{\mu}_n, \mathbf{\Sigma}_n)$ from the closed-form Gaussian case in Eqs.~\eqref{eq:GaussianPosteriorMean}, \eqref{eq:GaussianPosteriorCov}

\begin{align}
 \hat{\boldsymbol{\mu}}_n&
    \;:=\; \frac{1}{N_s}\sum_{i=1}^{N_s} \mathbf{x}^{(i)},\label{eq:StateApproximationMean}\\
    \hat{\mathbf{\Sigma}}_n &\;:=\; \frac{1}{N_s - 1}\sum_{i=1}^{N_s}
    \bigl(\mathbf{x}^{(i)} - \hat{\boldsymbol{\mu}}_n\bigr)
    \bigl(\mathbf{x}^{(i)} - \hat{\boldsymbol{\mu}}_n\bigr)^\top.
    \label{eq:StateApproximation}
\end{align}

which compresses the posterior into a low-dimensional pair of Gaussian moments that can be processed efficiently by the policy network. We study the numerical impact of the sample size $N_s$ on policy learning in Sec.~\ref{sec:NumericalStudies} for the one-active-pixel prior.

\subsection{Reward computation and approximation: Noiseless and noisy cases}\label{sec:RewardApprox}

The EIG reward $R_n = \mathrm{EIG}_{\mathbf{x}|\mathbf{y}_{n-1};\mathbf{H}_{n-1}}(\mathbf{h}_n)$ is evaluated at each step of every training trajectory. Computing it exactly requires integrating over both the unknown image $\mathbf{x}$ and the future observation $y_n$, which is tractable only under specific model assumptions. Two variants appear as the posterior $P(\mathbf{x}\mid\mathbf{y}_{n-1};\mathbf{H}_{n-1})$ and the likelihood $P(y_n\mid\mathbf{x};\mathbf{h}_n)$ admit closed-form representations or not.

For a general prior $P(\mathbf{x})$ and likelihood $P(y_n\mid\mathbf{x};\mathbf{h}_n)$, the exact reward is

\begin{align}\label{eq:EIG-general}
    \begin{split}
 R_n &= \mathrm{EIG}_{\mathbf{x}|\mathbf{y}_{n-1};\mathbf{H}_{n-1}}(\mathbf{h}_n)
     \\&= \mathrm{Ent}\bigl[P(\mathbf{x}\mid\mathbf{y}_{n-1};\mathbf{H}_{n-1})\bigr]
     - \mathbb{E}_{y_n}\!\Bigl[\mathrm{Ent}\bigl[P(\mathbf{x}\mid y_n,\mathbf{y}_{n-1};\mathbf{H}_{n})\bigr]\Bigr],       
    \end{split}
\end{align}

where the expectation is over the marginal $P(y_n;\mathbf{h}_n) = \int_\mathcal{X} P(y_n\mid\mathbf{x};\mathbf{h}_n)\,P(\mathbf{x}\mid\mathbf{y}_{n-1};\mathbf{H}_{n-1})\,\mathrm{d}\mathbf{x}$. In general, evaluating~\eqref{eq:EIG-general} exactly is intractable because it requires integrating over both $\mathbf{x}$ and $y_n$.

Given $N_{\mathrm{r}}$ posterior samples $\{\mathbf{x}^{(i)}\}_{i=1}^{N_{\mathrm{r}}}\!\sim P(\mathbf{x}\mid\mathbf{y}_{n-1};\mathbf{H}_{n-1})$ one can approximate the marginal as

\begin{align}\label{eq:marginal-MC}
 \hat{P}(y_n;\mathbf{h}_n) = \frac{1}{N_{\mathrm{r}}}\sum_{i=1}^{N_{\mathrm{r}}} P\!\left(y_n\mid\mathbf{x}^{(i)};\mathbf{h}_n\right).
\end{align}

To estimate the EIG itself, a single held-out sample is drawn $\mathbf{x}^{(0)}\sim P(\mathbf{x}\mid\mathbf{y}_{n-1};\mathbf{H}_{n-1})$, independent of the $N_{\mathrm{r}}$ denominator particles, and use the nested Monte Carlo (NMC) estimator~\cite{vincent2017darc}

\begin{align}\label{eq:EIG-MC}
 \hat{R}_n = \int_\mathcal{Y} P(y_n\mid\mathbf{x}^{(0)};\mathbf{h}_n) \left[\log P(y_n\mid\mathbf{x}^{(0)};\mathbf{h}_n)
           - \log \hat{P}(y_n;\mathbf{h}_n)\right] dy_n.
\end{align}

The one-dimensional integral in~\eqref{eq:EIG-MC} is over the observation space $\mathcal{Y}$ and can be evaluated analytically or by quadrature depending on the form of the likelihood.

In in absence of noise, i.e.\ $y_n = \langle\mathbf{h}_n, \mathbf{x}\rangle$, the likelihood collapses to a point mass $P(y_n\mid\mathbf{x};\mathbf{h}_n) = \delta\bigl(y_n - \langle\mathbf{h}_n, \mathbf{x}\rangle\bigr)$. In this case the integral over $\mathcal{Y}$ in~\eqref{eq:EIG-MC} vanishes and the estimator reduces to a purely discrete sum over particles. Concretely, the forward model is evaluated at the held-out sample to obtain $y_n^{(0)} = \langle\mathbf{h}_n, \mathbf{x}^{(0)}\rangle$ and at each denominator particle to form the empirical marginal likelihood. Because $y_n$ takes values in a finite set $\{\langle\mathbf{h}_n, \mathbf{x}^{(i)}\rangle\}_{i=0}^{N_\mathrm{r}}$, the EIG estimate becomes

\begin{align}\label{eq:EIG-noiseless}
 \hat{R}_n^{\text{noiseless}} = -\log\!\left(\frac{1}{N_{\mathrm{r}}}\sum_{i=1}^{N_{\mathrm{r}}} \mathbf{1}\!\left[\langle\mathbf{h}_n, \mathbf{x}^{(i)}\rangle = \langle\mathbf{h}_n, \mathbf{x}^{(0)}\rangle\right]\right),
\end{align}

where $\mathbf{1}[\cdot]$ is the indicator function. This is simply the negative log of the fraction of particles that predict the same observation as the held-out sample.

When the observation is corrupted by additive uncorrelated Gaussian noise, $y_n = \langle\mathbf{h}_n, \mathbf{x}\rangle + \varepsilon$ with $\varepsilon\sim\mathcal{N}(0,\sigma^2)$, the likelihood becomes $P(y_n\mid\mathbf{x};\mathbf{h}_n) = \mathcal{N}\!\left( \langle\mathbf{h}_n, \mathbf{x}\rangle,\,\sigma^2\right)$. One can approximate it via numerical quadrature on a uniform grid $\{y_k\}_{k=1}^{K}$ spanning $[\min_i\langle\mathbf{h}_n,\mathbf{x}^{(i)}\rangle - 5\sigma,\; \max_i\langle\mathbf{h}_n,\mathbf{x}^{(i)}\rangle + 5\sigma]$ with spacing $\Delta y$. This quadrature scheme can be viewed as a deterministic variant of the nested Monte Carlo (NMC) estimator of~\cite{huan2013simulation}, where the outer expectation over observations is resolved on a fixed grid rather than by sampling; this removes the variance of the outer loop at the cost of scaling linearly with the grid resolution $K$, which remains tractable when the observation space is one-dimensional.

At each grid point $y_k$, define the weighted likelihood as

\begin{align}
    q_{ik} = P(\mathbf{x}^{(i)} \mid \mathbf{y}_{n-1}; \mathbf{H}_{n-1}) \cdot \mathcal{N}\!\left(y_k;\, \langle\mathbf{h}_n, \mathbf{x}^{(i)}\rangle,\, \sigma^2\right),
\end{align}

the marginal $m_k = \sum_{i=1}^{N_{\mathrm{x}}} q_{ik}$, and the posterior weights:

\begin{align}
    P(\mathbf{x}^{(i)} \mid y_k, \mathbf{y}_{n-1}; \mathbf{H}_{n-1}, \mathbf{h}_n) = \frac{q_{ik}}{m_k},
\end{align}

where $\cdot$ is the scalar product. 

The expected posterior entropy is obtained by averaging the entropy of the posterior distribution over the observation space via quadrature:

\begin{align}\label{eq:exp-post-ent-gaussian}
    \hat{\mathbb{E}}_{y_n}[\mathrm{Ent}\bigl[P(\mathbf{x}\mid y_n,\mathbf{y}_{n-1};\mathbf{H}_{n})]] &= \sum_{k=1}^{K} m_k \;\mathrm{Ent}[P(\mathbf{x} \mid y_k, \mathbf{y}_{n-1}; \mathbf{H}_{n-1}, \mathbf{h}_n)]\; \Delta y,
\end{align}

where $m_k$ acts as the quadrature weight, combining the marginal density with the grid spacing. The expected information gain is then

\begin{align}\label{eq:EIG-gaussian}
    \hat{R}_n^{\text{Gaussian}} = \mathrm{Ent}[P(\mathbf{x} \mid\mathbf{y}_{n-1}; \mathbf{H}_{n-1})] - \hat{\mathbb{E}}_{y_n}[\mathrm{Ent}\bigl[P(\mathbf{x}\mid y_n,\mathbf{y}_{n-1};\mathbf{H}_{n})]].
\end{align}

\subsubsection{Gaussian Prior case: D-Optimal Sequential Criterion}
When the prior is Gaussian, $P(\mathbf{x}) = \mathcal{N}(\mathbf{x}; \boldsymbol{\mu}_0, \boldsymbol{\Sigma}_0)$, the forward operator is linear and the noise is Gaussian, the posterior remains Gaussian at every step $n$ with mean $\boldsymbol{\mu}_{n-1}$ and covariance $\boldsymbol{\Sigma}_{n-1}$ updated via the recursions in ~\eqref{eq:GaussianPosteriorCov} and ~\eqref{eq:GaussianPosteriorMean}. In this setting, the model evidence for the $n$-th observation is itself Gaussian, $P(y_n \mid \mathbf{h}_n) = \mathcal{N}(y_n;\, \mathbf{h}_n^\top \boldsymbol{\mu}_{n-1},\, s_n^2)$, with predictive variance $s_n^2 = \mathbf{h}_n^\top \boldsymbol{\Sigma}_{n-1} \mathbf{h}_n + \sigma^2$. Since the posterior covariance $\boldsymbol{\Sigma}_n$ is independent of the observed value $y_n$, averaging over the model evidence via Gaussian moment identities yields a closed-form reward

\begin{equation}
    R_n^{\mathrm{Gauss}}(\mathbf{h}_n)
    = \frac{1}{2} \log \det\!\left(
        \mathbf{I} + \sigma^{-2}\,\boldsymbol{\Sigma}_{n-1}^{1/2}\,
        \mathbf{h}_n \mathbf{h}_n^\top\,
        \boldsymbol{\Sigma}_{n-1}^{1/2}
      \right)
    = \frac{1}{2} \log\!\left(
        1 + \frac{\mathbf{h}_n^\top \boldsymbol{\Sigma}_{n-1} \mathbf{h}_n}{\sigma^2}
      \right),
    \label{eq:gauss_reward}
\end{equation}
where the second equality follows from the matrix determinant lemma~\cite{HornJohnson2013}.

This is precisely the \emph{D-optimal design criterion} evaluated sequentially~\cite{chaloner1995bayesian, fedorov1972theory}: the optimal pattern $\mathbf{h}_n^*$ at step $n$ maximises the predictive variance $\mathbf{h}_n^\top \boldsymbol{\Sigma}_{n-1} \mathbf{h}_n$ of the current Gaussian belief, automatically adapting to all past measurements through the running covariance $\boldsymbol{\Sigma}_{n-1}$. The total EIG over $N$ steps then decomposes additively as 
\begin{equation}
    \mathrm{EIG}(\mathbf{H}_N)
    = \sum_{n=1}^{N} R_n^{\mathrm{Gauss}}(\mathbf{h}_n)
    = \frac{1}{2} \log \det\!\left(
        \mathbf{I} + \sigma^{-2}\, \boldsymbol{\Sigma}_0^{1/2}
        \mathbf{H}_N^\top \mathbf{H}_N
        \boldsymbol{\Sigma}_0^{1/2}
      \right),
    \label{eq:EIG_gaussian_D_optimal}
\end{equation}
recovering the standard Bayesian D-optimality objective~\cite{chaloner1995bayesian}.

Unlike the general estimator in~\eqref{eq:EIG-gaussian}, \eqref{eq:EIG_gaussian_D_optimal} requires no outer quadrature over $\mathcal{Y}$ and no posterior samples, only the covariance update, making it exact and computationally negligible.

\subsection{Pattern design including budget constraints}
\label{sec:BudgetConstraint}

The framework developed so far optimises binary measurement patterns to maximise information gain, without placing any other explicit restrictions on them. In many real-world applications, however, sensing is not free: every observation consumes a physical resource, as time, energy, or radiation dose, whose accumulation can harm the very system being studied. In live-cell fluorescence microscopy, for instance, cumulative light exposure directly damages the cells through multiple photodamage mechanisms~\cite{laissue2017assessing,icha2017phototoxicity}: excitation light generates reactive oxygen species that degrade cellular components, induces phototoxic stress responses altering gene expression, and accelerates photobleaching of fluorophores. The severity of these effects scales with total photon dose, which is simply the $\ell_{1,1}$ norm of $\mathbf{H}_N$. The dose-constrained sequential optimal experimental design for single-pixel imaging is defined as

\begin{equation}\label{eq:budget_objective}
			\pi_B^* := \argmax_{\pi} \; J(\pi) \quad \text{ s.t.} \quad \sum_{i,j} \left( \mathbf{H}_N \right)_{i,j} \leq B 		\notag
\end{equation}
The budget $B$ encourages the policy $\pi_B^*$ to focus sensing capacity on the most informative regions, naturally promoting sparse, adaptive measurement strategies. 
 
One approach to enforce the hard constraint is to directly optimize over policies that fulfill it, e.g., by using the conditional Bernoulli policies $\pi^{CBer}_{\theta,B}(A_n \mid S_{n-1})$~\cite{attia2024probabilistic}, which generates each new pattern conditioned on the dose used so far. While using these policies guarantees strict dose compliance, it introduces sequential dependencies across pattern proposals, which can slow training and requires the policy to anticipate future sensing opportunities.

An alternative is to fold the cost into the objective as a soft constraint, penalizing dose expenditure directly within the policy gradient
\begin{equation}\label{eq:soft_constraint}
			\pi_\lambda^* := \argmax_{\pi} \; J(\pi) - \lambda \sum_{i,j} \left( \mathbf{H}_N \right)_{i,j} 
\end{equation}

where $\lambda>0$ balances information gain against dosage used. This Lagrangian relaxation allows more flexible pattern proposals and avoids the sequential conditioning required by the hard-constrained approach, but does not guarantee strict dose compliance: A too small $\lambda$ may lead to $B$ routinely exceeded; too large $\lambda$ and informative patterns may be suppressed. 

\subsection{Learning experimental designs using parametric policies}\label{sec:ParametricPolicies}

For policy gradient methods, improving the experimental design reduces to iteratively updating the policy parameters $\theta$ introduced in Sec. \ref{sec:PolicyArchitectureComparison} to maximize the expected information gain objective $J(\pi_\theta) = J(\theta)$ by gradient ascent

\begin{equation}\label{eq:GradientAscent}
    \theta_{n} = \theta_{n-1} + \alpha\nabla_{\theta}J(\theta_n)
\end{equation}

where $\alpha$ is learning rate of the parameterization update. 

As computing the true gradient is intractable, different methods rely on different stochastic approximation approaches. When a differentiable policy is available, one can make use of the gradient policy theorem \cite{sutton1998temporal}, to reformulate the gradient of ~\eqref{eq:RL-sOED} as:

\begin{align}\label{eq:PG-sOED}
\begin{split}
            \nabla_\theta J(\theta)  &= \nabla_\theta \mathbb{E}_{\tau\sim P(\tau \mid \pi_{\theta})}\Bigg[\sum_{n=1}^{N} R_n(S_{n-1}, A_{n}, S_{n})\Bigg]\\
&= \mathbb{E}_{\tau \sim P(\tau \mid \pi_{\theta})}\Bigg[\sum_{n=1}^{N}G_n \nabla_{\theta}\log \pi_{\theta}(A_n\mid S_{n-1})\Bigg]\ \quad , 
    \end{split}
\end{align}

where $G_{n} := \sum_{l=n}^{N} R_l$ is the expected cumulative reward.

At each step $n$, the update is proportional to $G_n = \sum_{l=n}^N R_l$, the sum of all rewards collected from that step until the end of the experiment. This means the policy is geared towards choices that proved beneficial across the remainder of the episode. Actions that consistently produce rewards are reinforced, while those that led to diminishing returns are gradually discarded. By doing this, the expected cumulative return attributes long-term success back to the decisions that were crucial in the past.

The policy gradient update is the core step in the REINFORCE algorithm \cite{SuBa2018}, which approximates the expected value in \eqref{eq:PG-sOED} by Monte Carlo estimation over $N_{MC}$ trajectory samples: For each trajectory sample, the expected cumulative reward is computed for each step from that point until $n=N$ and using \eqref{eq:PG-sOED} the gradient in \eqref{eq:GradientAscent} is approximated as

\begin{equation}\label{eq:MC-Gradient-Ascent}
\nabla_\theta J(\theta)\ \approx \frac{1}{N_{MC}}\sum^{N_{MC}}_{j=1} \sum_{n=1}^{N} G^{(j)}_n \nabla_\theta \log \pi_\theta(A^{(j)}_n\mid S^{(j)}_{n-1})    
\end{equation}

The full REINFORCE algorithm formulated for the sOED problem is described in Algorithm \ref{alg:REINFORCESPI}.

\begin{algorithm}[H]
\caption{REINFORCE for SPI Sequential Optimal Experimental Design}
\label{alg:REINFORCESPI}
\begin{algorithmic}[1]
\State \textbf{Initialize} policy parameters $\theta$ randomly
\For{episode $i = 1$ to $N_{\text{episodes}}$}
\State sample ground truth image $\mathbf{x} \sim P(\mathbf{x})$ for this episode
\State //~Collect $N_{MC}$ trajectories
\For{trajectory $j = 1$ to $N_{MC}$}
\State $\tau^{(j)} \leftarrow \emptyset$
\State $S_0^{(j)} \leftarrow P(\mathbf{x})$ (prior distribution evaluated at $\mathbf{x}$)
\For{step $n = 1$ to $N$}
\State $\mathbf{h}_n^{(j)} \sim \pi_\theta(\cdot \mid S_{n-1}^{(j)})$
\State $y_n^{(j)} \sim P(y \mid \mathbf{x};\mathbf{h}_n^{(j)})$ 
\State $R_n^{(j)} = \mathrm{EIG}_{\mathbf{x}\mid\mathbf{y}_{n-1}^{(j)};\mathbf{H}_{n-1}^{(j)}}(\mathbf{h}_n^{(j)})$ as described in  Sec. \ref{sec:RewardApprox}
\State $S_n^{(j)} \leftarrow P(\mathbf{x} \mid \mathbf{y}_n^{(j)};\mathbf{H}_n^{(j)}) \propto P(y_n^{(j)} \mid \mathbf{x};\mathbf{h}_n^{(j)}) \cdot S_{n-1}^{(j)}$ as described in Sec. \ref{sec:StateApprox}
\State $\tau^{(j)} \leftarrow \tau^{(j)} \cup \{(S_{n-1}^{(j)},\mathbf{h}_n^{(j)},R_n^{(j)}, S_n^{(j)})\}$ as described in Sec. \ref{sec:RL-SPI-sOED}
\EndFor
\State
\State// Compute expected cumulative reward
\For{step $n = 1$ to $N$}
\State $G_n^{(j)} \leftarrow \sum_{l=n}^N R_l^{(j)}$ (sum of rewards from step $n$)
\EndFor
\EndFor
\State 
\State //Update policy using \eqref{eq:MC-Gradient-Ascent}
\State $\nabla_{\theta}J(\theta) \leftarrow \frac{1}{N_{MC}} \sum_{j=1}^{N_{MC}} \sum_{n=1}^N  G_n^{(j)} \cdot \nabla_\theta \log \pi_\theta(A^{(j)}_n = \mathbf{h}_n^{(j)} \mid S_{n-1}^{(j)})$
\State Update parameters $\theta \leftarrow \theta + \alpha \cdot \nabla_{\theta}J(\theta)$ as in \eqref{eq:GradientAscent}
\EndFor
\State \Return Trained policy $\pi_{\theta}$
\end{algorithmic}
\end{algorithm}

\section{Numerical Studies}\label{sec:NumericalStudies}

The proposed reinforcement learning approach for sOED is validated by learning optimal measurement strategies in a simple, yet representative, numerical SPI scenario. Concepts introduced in the previous section and their impact on learning performance, such as the policy parametrization (cf. Sec, \ref{sec:ParametricPolicies}), state approximation (cf. Sec \ref{sec:StateApprox}), reward approximation (cf. Sec \ref{sec:RewardApprox}), dose constraints (cf. Sec. \ref{sec:BudgetConstraint}) performance are illustrated. The ability of the framework to find optimal strategies under increasing measurement noise is shown.

\subsection{One-active-pixel SPI scenario}\label{sec:OAPScenario}

\begin{figure}
    \centering \includegraphics[width=1\linewidth]{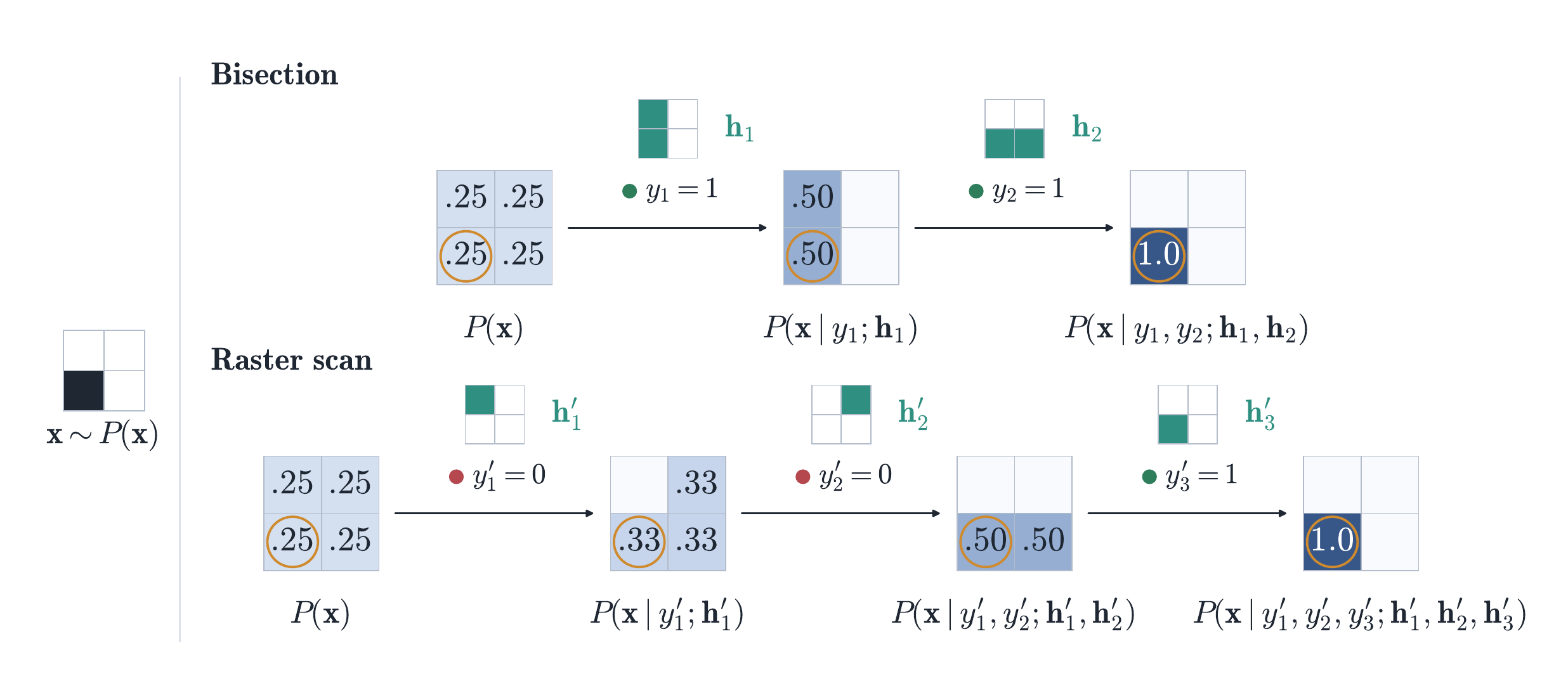}
    \caption{The posterior is being updated using measurements of the ground truth $\mathbf{x}$ via the proposed patterns $\mathbf{h}_n$ (in green). Bisection and raster scan are considered as two possible strategies to identify the ground truth from measurements via one-active-pixel prior knowledge.}
    \label{fig:OAP_Trajectory}
\end{figure}

Consider a structured single-pixel imaging setup in which exactly one pixel of the unknown image is active with intensity $1$, as described in the left side of Figure \ref{fig:OAP_Trajectory}. This discrete sparsity-constraint means that the possible images are reduced to the discrete set of canonical basis vectors, and one can assume that the prior $P(\mathbf{x})$ accurately captures this a-priori information. In this setting, both the Bayesian posterior and the expected information gain are tractable to compute either in closed form, for the noiseless scenario, or via low-dimensional numerical quadrature in the noisy case. It provides a simple benchmark in which the reinforcement learning agent can use the exact posterior as the state and the sensing strategy can be compared against analytically optimal designs as well as against random sensing pattern. It also enables a systematic study of variability across independent policy initializations, since the discrete prior supports repeated runs with a modest computational cost.

Formally, the one-active-pixel set can be described by the images   
$\mathcal{E}_M := \{\mathbf{e}_1,\dots,\mathbf{e}_M \},
$
where \(\mathbf{e}_m\) denotes the $m$-th unit vector in $\mathbb{R}^M$. The prior $P(\mathbf{x})$ is a discrete probability measure on $\mathbb{R}^M$ and can be written as

\begin{equation}\label{eq:OAP-prior}
    P(\mathbf{x}) \;=\; \sum_{m=1}^M w^{(0)}_m \, \delta_{\mathbf{e}_m}(\mathbf{x}),
    \qquad w^{(0)}_m = 1/M
\end{equation}

where $\delta_{\mathbf{e}_m}(\mathbf{x})$ denotes a unit point mass concentrated at $\mathbf{x}=\mathbf{e}_m$. Following \eqref{eq:Bayes}, each measurement leads to posterior that is also discrete and has the same form as the prior, just with updated weights

\begin{equation}\label{eq:OAP-posterior}
    P(\mathbf{x} \mid \mathbf{y}_n; \mathbf{H}_n)
    \;=\;
    \sum_{m=1}^M w^{(n)}_m \, \delta_{\mathbf{e}_m}(\mathbf{x}), \quad w^{(n)}_m \propto P(\mathbf{y}_n \mid \mathbf{x}=\mathbf{e}_m; \mathbf{H}_n), \quad \sum_{m=1}^M w^{(n)}_m = 1
\end{equation}

Thus, the posterior and the RL state can be fully represented by the $M$-dimensional vector $S_n = \mathbf{w}^{(n)} = (w^{(n)}_1 ,\ldots,w^{(n)}_M)$. 

In the ideal noiseless one-active-pixel setting, a measurement of any $\mathbf{x} = \mathbf{e}_m$ that is supported by the prior with a  binary measurement pattern $\mathbf{h} \in \{0,1\}^M $ yields either  $0$ or $1$

\begin{equation}\label{eq:likelihood}
    y = \langle \mathbf{h}, \mathbf{e}_m \rangle = \begin{cases}
    1  & \textrm{if} \qquad (\mathbf{h})_m = 1 \\
    0  & \textrm{otherwise} 
    \end{cases}
\end{equation}

If one measures $y = 1$, the active pixel then lies in the support of the illuminated mask, i.e., $m \in \textrm{supp}(\mathbf{h})$. Updating the prior with this likelihood information means removing all one-active-pixel images with active locations outside of the mask and renormalizing the weights to yield a uniform distribution over the remaining locations, as illustrated in the trajectories of Figure \ref{fig:OAP_Trajectory}. Measuring $y=0$ accordingly removes all locations that are in the support of the illuminated mask from the posterior. Intuitively, this scenario thereby corresponds to a simple "finding a needle in a haystack" search problem: where, out of $m$ possible locations, a particular objects sits. Every measurement one can make with a pattern $\mathbf{h}$ corresponds to asking the question "Is the object located in any of the illuminated positions?" and receiving a "yes-or-no" answer.

For illustration, let us consider the case $M = 4$ case: The prior is uniform over $\{\mathbf{e}_1,\dots,\mathbf{e}_4\}$, leading to $\mathbf{w}^{(0)} = (1/4, 1/4, 1/4, 1/4)$. The simplest case is the conventional raster scan, where one pixel at a time is illuminated in a fixed order. The sequence of patterns would be $\mathbf{h}_1 = (1,0,0,0)$, $\mathbf{h}_2 = (0,1,0,0)$, $\mathbf{h}_3 = (0,0,1,0)$, $\mathbf{h}_4 = (0,0,0,1)$. Now suppose that the true image is $\mathbf{x} = \mathbf{e}_3$. The first measurement is then $y_1 = 0$, which discards $\mathbf{e}_1$ but leaves $\mathbf{e}_2,\mathbf{e}_3,\mathbf{e}_4$ as possible. After the likelihood update, the posterior weights become

\begin{equation}
    \mathbf{w}^{(1)} = \Bigl(0,\,\tfrac{1}{3},\,\tfrac{1}{3},\,\tfrac{1}{3}\Bigr).
\end{equation}

The second pattern yields again $y_2 = 0$, ruling out $\mathbf{e}_2$ and leaving only $\mathbf{e}_3$ and $\mathbf{e}_4$ feasible, so the posterior weights are

\begin{equation}
    \mathbf{w}^{(2)} = \Bigl(0,\,0,\,\tfrac{1}{2},\,\tfrac{1}{2}\Bigr).
\end{equation}

Only after the third measurement, with $\mathbf{h}_3 = (0,0,1,0)$ and outcome $y_3 = 1$, does the posterior collapses to a single delta distribution centered on the true image, such that weights are

\begin{equation}
    \mathbf{w}^{(3)} = (0,\,0,\,1,\,0).    
\end{equation}

In this worst-case trajectory, the posterior support shrinks from four to three, then to two, and only then to a single element, illustrating that, under a uniform prior, raster scanning requires on average $(M+1)/2$ measurements to uniquely identify the active pixel, exhibiting a linear measurement complexity in the number of image elements $M$.

In analogy to search problems, a bisection strategy would be more advantageous, where the patterns are chosen to half the current possible locations: In the same $M=4$ example with uniform prior $\mathbf{w}^{(0)}$ and true image $\mathbf{x} = \mathbf{e}_3$, one can consider a first pattern $\mathbf{h}_1 = (1,1,0,0)$, which only illuminates pixels 1 and 2. The first outcome is then $y_1 = 0$ and after renormalizing the posterior becomes uniform over $\mathbf{e}_3$ and $\mathbf{e}_4$ so that the weights are 

\begin{equation}
    \mathbf{w}^{(1)} = \Bigl(0,\,0,\,\tfrac{1}{2},\,\tfrac{1}{2}\Bigr),
\end{equation}

An \emph{adaptive} second pattern is then $\mathbf{h}_2 = (0,0,1,0)$, which illuminates the third but not the fourth element. The second outcome is $y_2 = 1$, collapsing to a delta centered in $\mathbf{e}_3$ and the weights become 

\begin{equation}
    \mathbf{w}^{(2)} = (0,\,0,\,1,\,0),
\end{equation}

shrinking from four to two and then directly to a single element in just two measurements. 

Another possibility is the usage of a pattern that, on top of halving the possible locations, also keeps illuminating a partial fraction of the already measured pixels. From the previous case, a pattern $\mathbf{h}_2 = (0,1,1,0)$ represents a \emph{fixed} bisection strategy, and these strategies are different from the adaptive ones in the dose budget they require and their dependence on the measurements. Fixed bisection allows via redundancy to solve for all possible images in the one-active-pixel set without redesigning their activations given the measurements obtained over the experiment. 

For both cases, bisection strategies half the support of the posterior distribution at each step and, more generally, identifies the active pixel in at most $\lceil \log_2 M \rceil$ measurements, reflecting its logarithmic complexity in this simplified SPI setting.

\subsection{Optimal EIG strategies}\label{sec:OptimalEIGStrategies}

For the scenario described, one can find that bisection strategies are optimal with respect to EIG. Using this result, learned policies can be compared against the strategies yielding the best achievable performance. Firstly, the EIG \eqref{eq:EIG} is computed for images in the one-active-pixel set $x\in\mathcal{E} = \{\mathbf{e}_1, \ldots, \mathbf{e}_M\}$ for the first measurement with $\mathbf{h}$. The marginal likelihood $P(\mathbf{y};\mathbf{h})$ is needed to compute the expectation over the prior and posterior entropy difference. 

\begin{figure}
    \centering \includegraphics[width=1\linewidth]{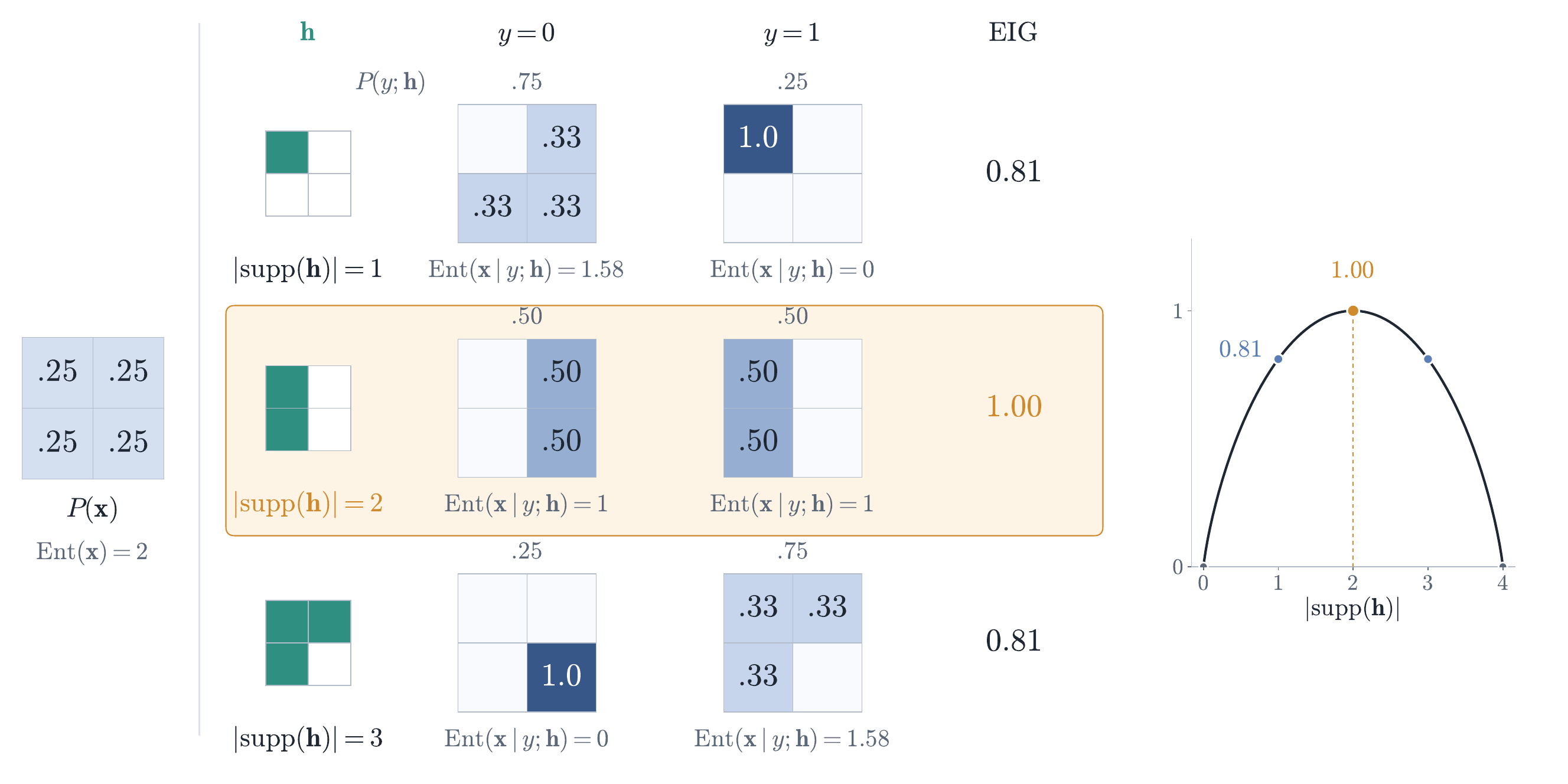}
    \caption{For a uniform one-active-pixel, strategies that illuminate either one or all but one site are suboptimal in an EIG sense. This can be seen from the unbalanced posterior at the different possible outcomes $\mathbf{y}$. For the bisection strategy, measurements 0 or 1 have equal probability and the posterior support halves, which produces the optimal one-step EIG.}
    \label{fig:OAP_OptimalStrategies}
\end{figure}

As illustrated in Figure \ref{fig:OAP_OptimalStrategies}, $y$ can only take the values $0$ and $1$ in the noiseless scenario: The marginal likelihood of $y = 1$ is the ratio between the number of illuminated sites in the measuring pattern $h$, $\mid \textrm{supp}(\mathbf{h}) \mid$ and the number of possible illuminated sites $M$, so that $P(y=1;\mathbf{h}) = \mid \textrm{supp}(\mathbf{h}) \mid/M$ and $P(y=0;\mathbf{h}) = (M - \mid \textrm{supp}(\mathbf{h}) \mid)/M$.

Then, the entropy for the uniform one-active-pixel prior, as described in \eqref{eq:OAP-prior}, reads

\begin{align}
\begin{split}
\mathrm{Ent}(\mathbf{x}) = \log M
\end{split}    
\end{align}

Following the previous section, the posterior shrinks with respect to the uniform prior. One can compute the posterior entropy by averaging over the possible images of the set after the measurement, which leaves 

\begin{align}
\begin{split}
    \mathrm{Ent}(\mathbf{x}\mid y; \mathbf{h})=& \begin{cases}
        \log (M - \mid \textrm{supp}(\mathbf{h})\mid) & if \qquad y = 0 \\
        \log (\mid \textrm{supp}(\mathbf{h})\mid) & if \qquad y = 1 
    \end{cases}
\end{split}
\end{align}

With this, one can write the one-step EIG as an information gain average weighted by the marginal likelihood $P(y;\mathbf{h})$. Decomposing the information gain as the difference between prior and posterior entropy, one obtains

\begin{align}\label{eq:EIG-OAP}
\begin{split}
    \mathrm{EIG}_{\mathbf{x}}(\mathbf{h}) =& \sum_{y = 0, 1} P(y;\mathbf{h})[\mathrm{Ent}(\mathbf{x}) - \mathrm{Ent}(\mathbf{x}\mid y; \mathbf{h})] \\
    =& \frac{\mid \textrm{supp}(\mathbf{h})\mid}{M}\log \frac{M}{\mid \textrm{supp}(\mathbf{h})\mid} + \frac{ M - \mid\textrm{supp}(\mathbf{h})\mid}{M}\log \frac{M}{M - \mid \textrm{supp}(\mathbf{h})\mid},
\end{split}
\end{align}
which means that $\mathrm{EIG}_{\mathbf{x}}(\mathbf{h})$ is a function of $s=\mid\textrm{supp}(\mathbf{h})\mid$. If $s$ was a continuous variable on $(0,M)$
\begin{equation}
  \frac{d}{ds}\,\mathrm{EIG}_{\mathbf{x}}(s) = \frac{1}{M}\log\frac{M-s}{s},
  \qquad
  \frac{d^2}{ds^2}\,\mathrm{EIG}_{\mathbf{x}}(s) = -\frac{1}{M}\Big(\frac{1}{s}+\frac{1}{M-s}\Big) < 0 ,
\end{equation}
so the $\mathrm{EIG}_{\mathbf{x}}(s)$ would be strictly concave w.r.t. $s$ and its unique stationary point
\begin{equation}
  s^\ast = \argmax_{s}\,\mathrm{EIG}_{\mathbf{x}}(s) = \frac{M}{2}
\end{equation}
is the global maximum. This means that for an even $M$, the optimal pattern illuminates half the sites of the possible image set $\mathcal{E}$ in one step.

Because every posterior is again uniform, now over the remaining image set $\mathcal{E}_i \subseteq \mathcal{E}$ of size $M_i$, with $M_0 = M$, the step-$(i{+}1)$ design problem is identical to the first with $M$ corresponding to $  M_i$ and $\mathrm{supp}(\mathbf{h}_{i+1}) \subseteq \mathcal{E}_i$. The same calculation gives $s_{i+1}^\ast = M_i/2$, and at the maximum both outcomes leave $M_i/2$ possible sites. Thereby, $M_{i+1} = M_i/2$ and the support halves at every step, leading to $M_i = M/2^i$. Bisection is then EIG-optimal per step and identifies the active pixel in exactly $\lceil \log_2 M\rceil$ measurements no matter where the active pixel is located.

\subsection{Experimental settings}
\label{sec:ExperimentalSettings}

In all presented RL experiments, $N=2$ is set as the number of measurements, so that the posterior is updated twice per experiment. Furthermore, $N_{MC} = 1$ is always used in \eqref{eq:MC-Gradient-Ascent}, i.e., the policy update is based on a single simulated trajectory. This corresponds to the setting one would use for large-scale imaging scenarios. Every RL experiment is repeated 100 times with different policy initializations to study the stability and convergence of learned designs and report the mean and $95\%$ confidence intervals in the evaluations. Unless stated otherwise, the total expected information gain of a policy $\pi_\theta$ is normalized for visualization purposes, to allow for the comparison of policies that propose binary pattern strategies under different experimental setups
\begin{equation}
    \mathrm{J}(\pi_\theta)
    \;=\;
    \frac{
        J({\pi_\theta})
        \;-\;
        \overline{\mathrm{J}}_{\mathrm{rand}}
    }{
        \mathrm{J}_{\mathrm{opt}}
        \;-\;
        \overline{\mathrm{J}}_{\mathrm{rand}}
    }\qquad,
    \label{eq:NIG}
\end{equation}

where the maximum EIG value $\mathrm{J}_{\mathrm{opt}}$ is calculated for that experiment, obtained using a bisection strategy in the one-active-pixel prior case, and the average EIG $\overline{\mathrm{J}}_{\mathrm{rand}}$ that a fully random policy reaches which illuminates each pixel with probability $0.5$.

\subsection{Policy architecture comparison}\label{sec:PolicyArchitectureComparisonExperiments}

\begin{figure}[tbp!]
    \centering
    \includegraphics[width=1\linewidth]{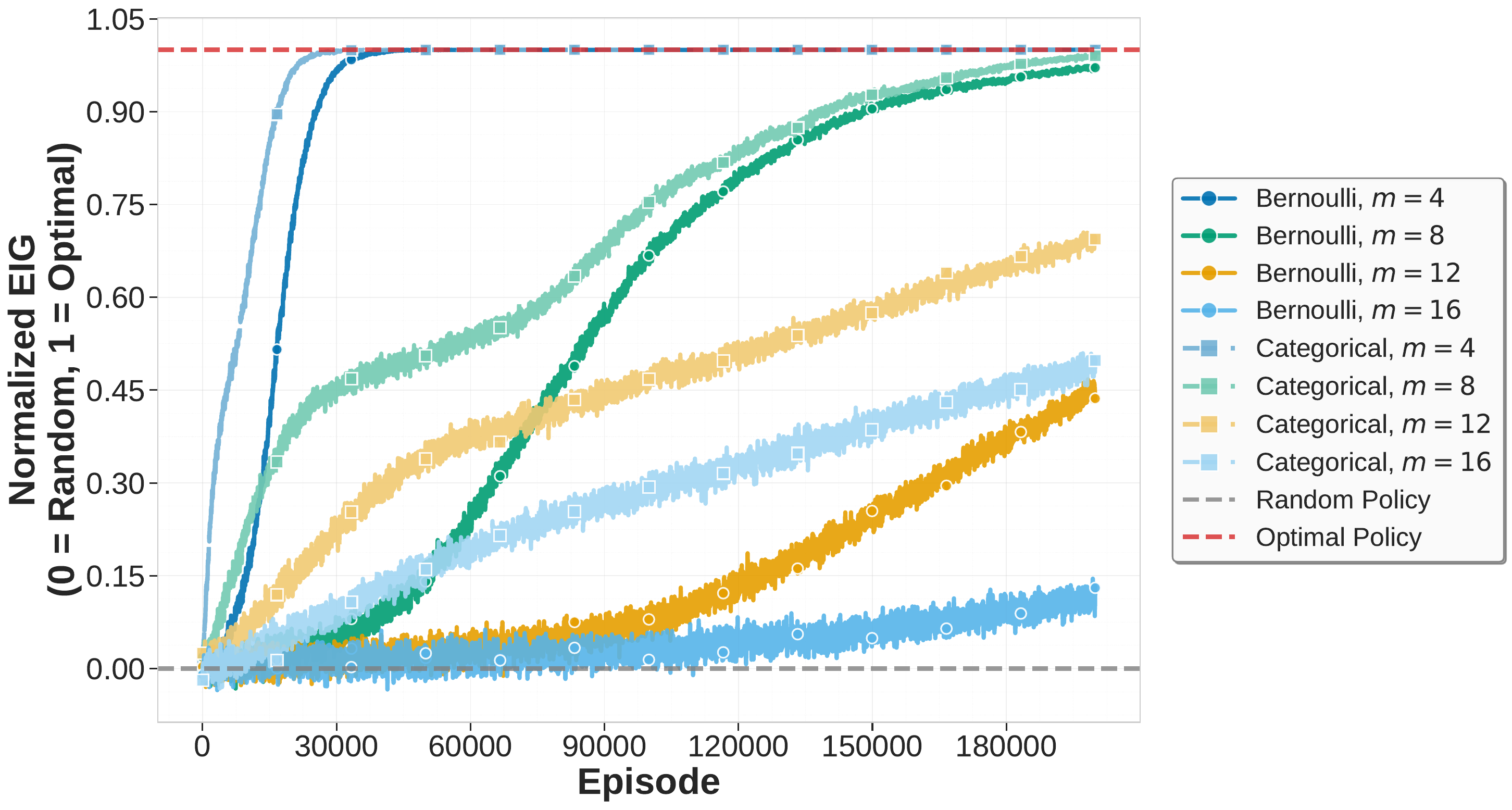}
    \caption{Total normalized EIG for categorical (lighter tone, square marker) and Bernoulli (darker tone, circle marker) policies trained over different image sizes $M = \{4, 8, 12, 16\}$ and same experimental step size $N = 2$. Optimal and random baselines are drawn as red and gray lines, respectively. }\label{fig:Results_NoiselessCategoricalBernoulliComparison}
\end{figure}

Figure \ref{fig:Results_NoiselessCategoricalBernoulliComparison} shows two training comparisons: the scaling of image sizes $M = \{4, 8, 12, 16\}$ and the usage of categorical or Bernoulli pattern proposal architectures, as described in Section \ref{sec:PolicyArchitectureComparison}. Categorical-based policies, described in \eqref{eq:CategoricalPolicy}, reach optimal convergence in fewer episodes than the Bernoulli-based ones, formulated in \eqref{eq:BernoulliPolicy}. Scaling the image size shows a higher requirement of episodes for convergence for the optimal EIG in two experimental steps.  

% Categorical-based policies as in \eqref{eq:CategoricalPolicy} reach convergence to the optimal performance for the smaller image size scenarios in fewer episodes than Bernoulli-based ones as formulated in \eqref{eq:BernoulliPolicy}. Large image sizes show a monotone increase in their performance, but the amount of episodes needed for convergence is still higher. The reward signal in the categorical case assigns each pattern their corresponding performance uniquely. The downside of this kind of architecture is their scalability: As the action space scales as $2^M$ in the output layer, their usage becomes computationally prohibitive as the image size grows. 

The normalized EIG at Figure \ref{fig:Results_NoiselessCategoricalBernoulliComparison} shows that Bernoulli policies are slower to train than categorical policies for all image sizes. However, Bernoulli policies can parameterize distributions over the action space with fewer trainable parameters, as the output of the network is the distribution over the pattern activations, and this output layer grows linearly with the image size $M$. Actions sampled from the Bernoulli policies are rewarded indirectly, since there is not an output unit-to-pattern correspondance as there is in the categorical. Then, the reward carries more stochasticity than in the case of categorical policies.  

% The rewards in this case are dense, since there is not a one-to-one correspondence between patterns and their probability distribution. For instance, a particular pattern that achieves a high EIG is rewarded by promoting a particular Bernoulli distribution evaluated at that action via the policy gradient theorem, but there could exist a different Bernoulli distribution from producing the same high-performing action. Hence, the Bernoulli policy that proposes an optimal strategy is such that, in mean value, designs adaptively for the SPI problem. 

In the categorical case, for an image size of $M=16$ the output layer has a size of $\sim65\mathrm{k}$ units given that the design space is $\{0,1\}^{16\times 16}$. Larger images result challenging for a categorical parameterization, leaving Bernoulli kind of policies for these cases despite their requirement for longer training times. The stochastic reward assignment, instead of a direct pattern-to-distribution reward, and the design space size are the trade-off variables for the framework for scaling up to large imaging problems.

% Over this scenario, there is not a clear sign of adaptiveness of the learned patterns, given that there exists no constraint over the dose budget. Despite this, there exists a inflection point for all cases around 0.5 of the normalized EIG, corresponding to learning a two-active-pixel bisection design over the first pattern. This inflection point is smoothened in the Bernoulli training case due to the previously described reward assignment .

\subsection{Effect of sample-based approximations}\label{sec:SampleBasedState}

We study sample-based approximations for state and reward, as introduced in Sections \ref{sec:StateApprox} and \ref{sec:RewardApprox}. This is important for translating this framework to more complex inference problems, where the posterior is accessible only through samples rather than in closed form. The focus of this section is to assess these effects separately, while in the Supplementary Data we present a brief study of both effects in combination with respect to the posterior without approximation.

\subsubsection{State approximation}

As described in Section \ref{sec:OAPScenario}, the posterior and RL state at time $n$ is completely described by the discrete parameter $\mathbf{w}^{(n)}$, whose entries $w^{n}_i$ contains the probability that the active pixel is at location $i$. Therefore, the empirical mean state approximation computed from $N_s$ posterior samples, as defined in \eqref{eq:StateApproximationMean}, converges to the true state $\mathbf{w}^{(n)}$. Here, we compare policies trained with this approximation against the exact posterior benchmark.

\begin{figure}[tbp!]
    \centering
    \includegraphics[width=0.8\linewidth]{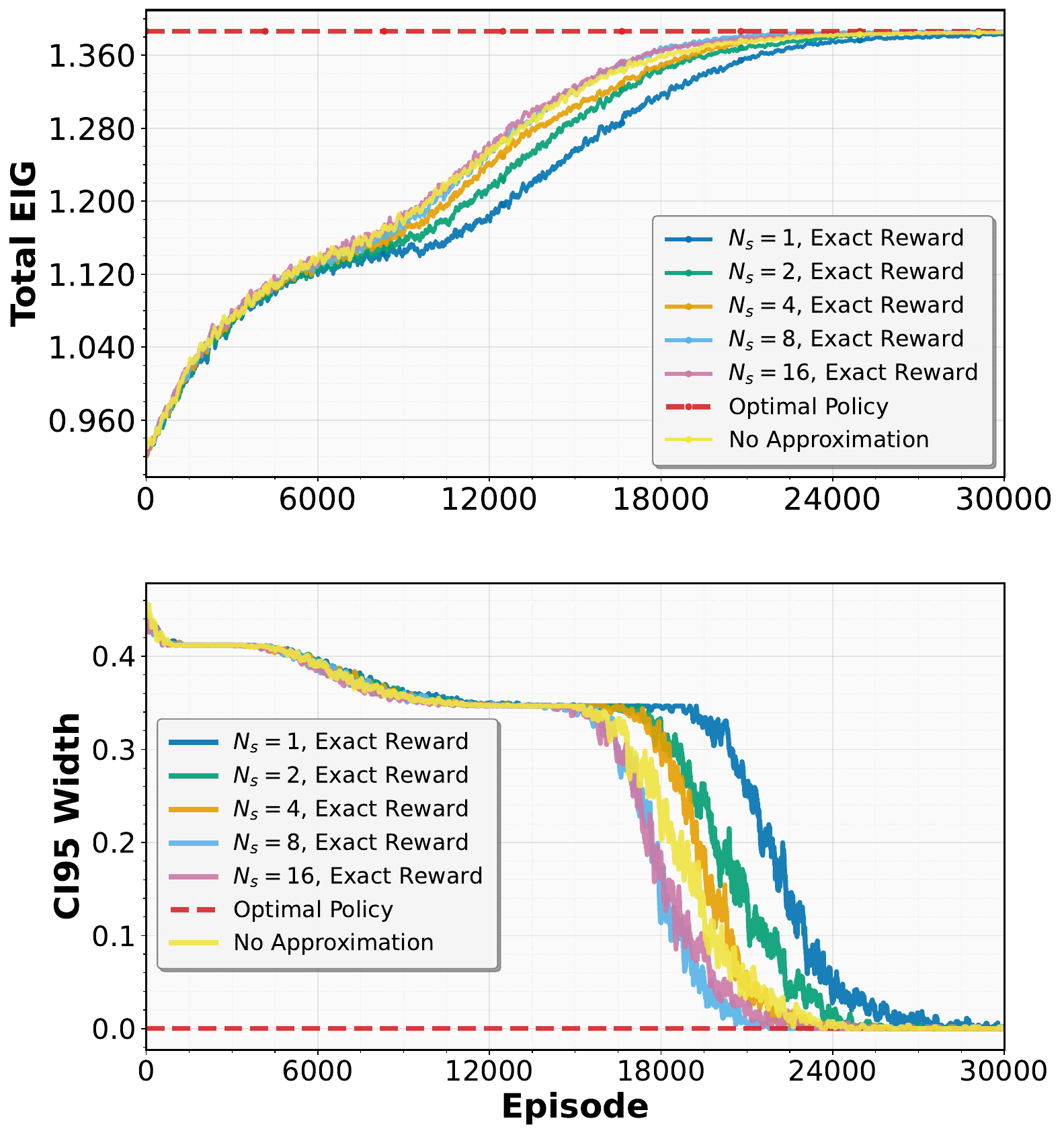}
    \caption{Total expected information gain for policy training under sample-based state approximations, with $N_s = \{1, 2, 4, 8, 16\}$. First row figure displays mean total EIG, second row shows $95\%$ confidence interval width. Curve for learning with no approximation, using exact state representation is illustrated in yellow for comparison of the sampling impact. Optimal policy reward is computed exactly and displayed in red for both plots.}    
    \label{fig:Results_StateApproximation_RewardExact}
\end{figure}

Figure \ref{fig:Results_StateApproximation_RewardExact} shows how approximating the state via sampling affects the learning process. Surprisingly, even the policy learned with approximating the state by a single sample reaches the optimal EIG. The inflection point around episode 6,000 marks the transition where modifying the number of samples produces a different learning performance. Strategies learned using fewer samples for the state approximation do need more episodes for reaching the optimal EIG.  The CI95 plots show that the variability follows a similar behavior: Policies trained with a low number of samples need more episodes to reduce the variability, but eventually all of them to to zero. 
%Given that the state update is detached from the policy update (cf. Algorithm \ref{alg:REINFORCESPI}), there is a computational tradeoff between the posterior sampling and the log-prob iteration, the latter usually involving a simple computation for the proposed architecture.

\subsubsection{Reward approximation}\label{sec:RewardApproximation}

We train policies to recover the optimal one-active-pixel strategy using the sample-based EIG estimator introduced in Section~\ref{sec:RewardApprox} as the reward signal. This setting allows us to assess whether policies trained with the estimator in \eqref{eq:EIG-MC} can recover the strategy that is optimal under the exact EIG objective in \eqref{eq:EIG-general}. Although evaluating the exact EIG requires the marginal likelihood, which is generally intractable in more complex settings, it is available in the scenario we designed and provides a benchmark for evaluation.

\begin{figure}[tbp!]
    \centering
    \includegraphics[width=0.8\linewidth]{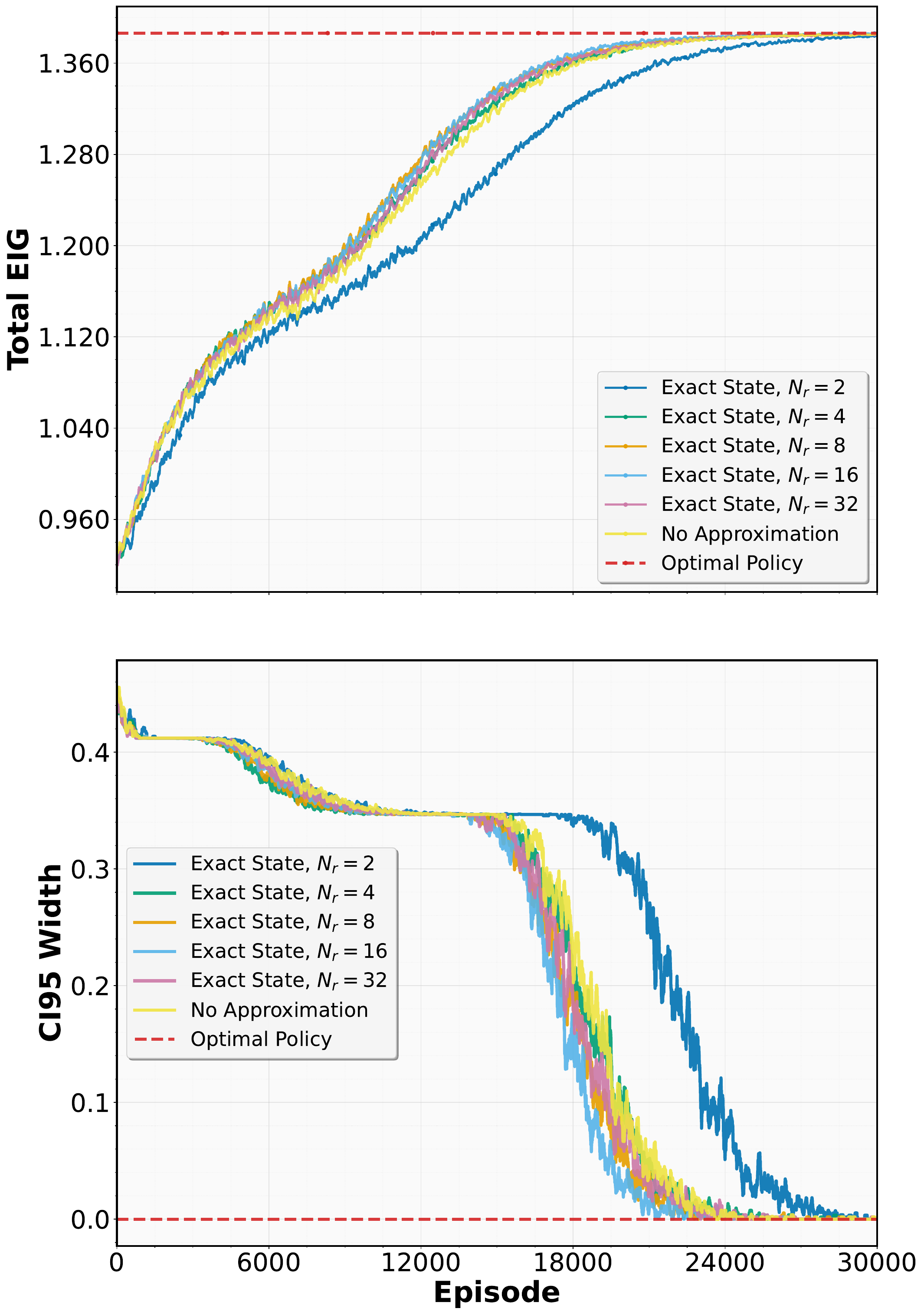}
    \caption{Total expected information gain using a sample-based reward approximation, with $N_r = \{2, 4, 8, 16, 32\}$ samples and a exact state representation. Optimal rewards for each sample quantity are illustrated in red, with mean values in the first row plot and $95\%$ confidence value widths over the second row. For comparison, EIG reward calculation using the posterior distribution representation without approximation ('No Approximation') is displayed in yellow.}
    \label{fig:Results_StateExact_RewardApproximation}
\end{figure}

In Figure \ref{fig:Results_StateExact_RewardApproximation} we observe that using the sample-based reward not only enables learning the optimal EIG strategy, it seems to even accelerate training. The only case displaying a slower training curve than the exact reward is the $N_r = 2$ case. In contrast to Figure \ref{fig:Results_StateApproximation_RewardExact}, after the inflection point in episode 6,000, using more than $N_r = 2$ samples for the reward approximation show a similar trend in mean total EIG and confidence interval widths to the case where the exact reward ('No Approximation') is used as a reward signal. Note that as in the sample-based state approximation (cf. Figure \ref{fig:Results_StateApproximation_RewardExact}), the training curves converge to the exact reward curves for larger numbers of samples. 

% As introduced in Section \ref{sec:RewardApprox}, Eq. \eqref{eq:EIG} presents a challenging computation due to the calculation of the marginal likelihood $P(\mathbf{y};\mathbf{H})$. If a posterior sampler is available, one can approximate the $\mathrm{EIG}$ computation by the NMC estimator presented in \eqref{eq:EIG-MC}, where the integral over measurements depends on the likelihood being used. For an exact state approximation, the NMC estimator and the exact $\mathrm{EIG}$ as learning metrics for a categorical policy are compared, computed as in \eqref{eq:CategoricalPolicy}. 

% In Figure \ref{fig:Results_StateExact_RewardApproximation}, different performance curves converge with a logarithmic scale to the exact $\mathrm{EIG}$ as more samples are added to the estimation. Since absolute $\mathrm{EIG}$ curves are being computed, this implies an exponential need of $N_r$ samples for approximating the exact expected information gain. 

\subsection{Learning under noisy measurements}\label{sec:NoisyMeasurements}

In the noiseless case, the first measurement collapses the posterior to delta distributions over the regions of activity or inactivity in the case of measuring a 1 or 0. In the noisy case $\sigma > 0$, this is no longer true, as noisy measurements still leave posterior probability mass over the discarded areas after the first acquisition. We now examine the impact of this on learning design policies. As a reference, we again compare their performance to the EIG of the bisection strategy, normalized as in \eqref{eq:NIG}. 

\begin{figure}[tbp!]
    \centering
    \includegraphics[width=0.9\linewidth]{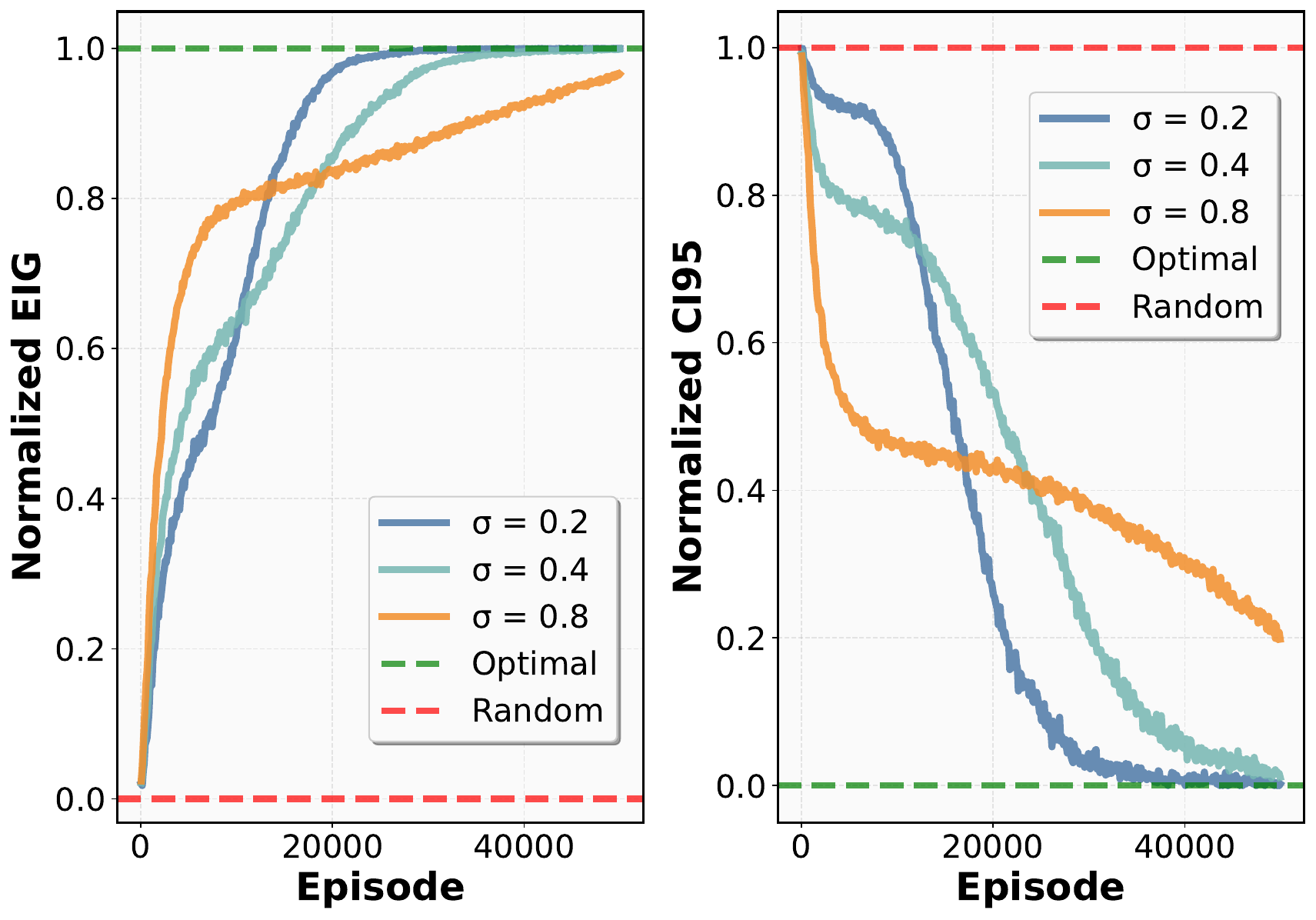}
    \caption{Total normalized EIG for three noisy measurement levels $\sigma \in \{0.2, 0.4, 0.8\}$. First column plot displays normalized EIG and second column plot $95\%$ confidence interval widths, averaged across 100 categorical policies realizations. Curves are normalized using random and optimal baselines for each noise level, and represented in red and green, respectively.}
    \label{fig:NoisyMeasurements}
\end{figure}

\begin{figure}[tbp!]
    \centering
    \includegraphics[width=1\linewidth]
    {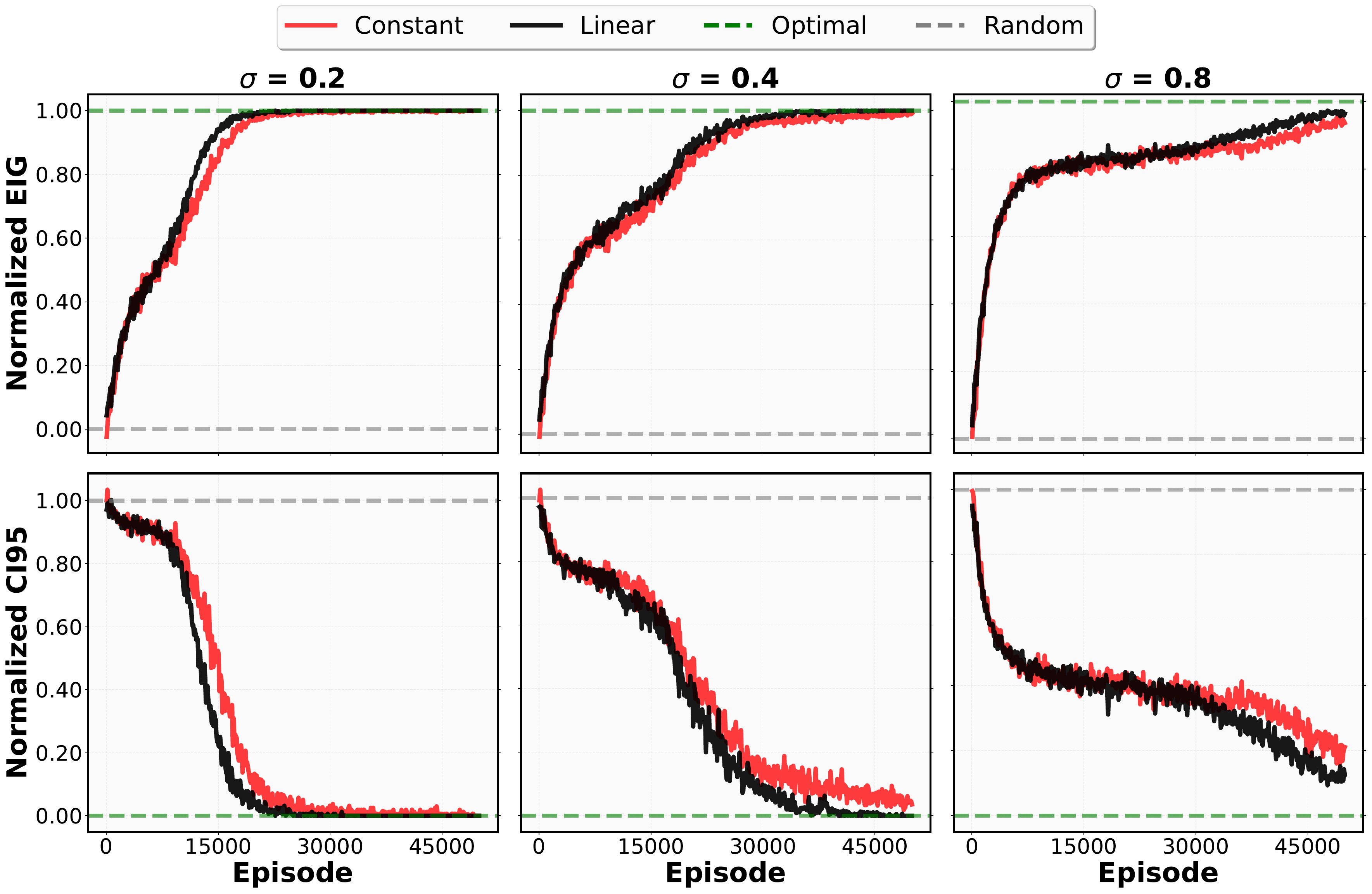}
    \caption{Total normalized EIG for three noisy measurement levels under constant and a linear curriculum learning for noise scheduling. Columns represent noise measurement levels $\sigma \in \{0.2, 0.4, 0.8\}$ and rows display the normalized means and $95\%$ confidence interval widths. Curves are normalized using random and optimal baselines for each noise level, and are drawn in gray and green lines, respectively.}
    \label{fig:Results_CurriculumLearning}
\end{figure}

Among the two approaches introduced in Section \ref{sec:RewardApprox} to treat the Gaussian likelihood modeled in \eqref{eq:SingleMeasurement}, our proposed one-active-pixel scenario requires the usage of a numerical integration method. For the experiments displayed in Figures \ref{fig:NoisyMeasurements} and \ref{fig:Results_CurriculumLearning}, the rewards using the Bayesian quadrature are computed as described in Section \ref{sec:RewardApprox}, with $K=1000$ elements over the grid determined by each sample. 

We can see that the learning behavior depends on the noise level: For a lower noise level, the policies learn slower than for the higher noise level at first but then overtake them and reach the reference normalized EIG  earlier (note however, that the normalization depends on $\sigma$). For higher noise levels, they reach a sub-optimal inflection point in less episodes, as we observe for $\sigma = 0.8$, but then it takes longer to converge to the reference EIG performance. 

% A episode-dependant linear noise schedule at the time of training $\sigma^{\mathrm{L}}_{\text{train}}(i) \;=\; \sigma_0 + (\sigma_{\text{goal}} - \sigma_0)\,\frac{i}{N_{\text{episodes}}}, \qquad i = 0, 1, \ldots, N_{\text{episodes}}$ is introduced. In this curriculum learning schema, the reward is calculated using $\sigma_{goal}$, and the learning with respect to a constant schedule $\sigma^{\mathrm{C}}_{\text{train}}(i) = \sigma_{\text{goal}}$ can be compared. The same configuration from the previous noisy measurement experiments is used.

Learning an optimal EIG strategy directly at a fixed measurement noise level $\sigma$ can be challenging, since noisy observations yield less informative posterior updates than in the noiseless setting. We investigate curriculum learning, in which training begins with noiseless measurements and the noise level is increased linearly to the target value $\sigma$ over the course of training. The episode-dependent noise schedule is $\sigma_{\text{train}}(i) = \sigma\,\frac{i}{N_{\text{episodes}}}$, for $i = 0, 1, \ldots, N_{\text{episodes}}$, where $i$ denotes the current episode. This curriculum is compared with the constant-noise training setting of Figure~\ref{fig:NoisyMeasurements}, in which $\sigma_{\text{train}}(i)=\sigma$ for every episode. To ensure a fair comparison, we evaluate both learned policies under the same target noise level $\sigma$ and compute their total EIG exactly using \eqref{eq:EIG-general}.

Using $\sigma \in \{0.2, 0.4, 0.8\}$, Figure \ref{fig:Results_CurriculumLearning} displays a faster learning of the linear schedule after the inflection point is reached around episode 15,000 for all noise levels. After the first episodes, the 95\% confidence intervals width shrink to zero for all $\sigma$, illustrating that using curriculum learning presents an performance advantage with respect to learning under constant noise. The linear curriculum learning policy is trained with cleaner measurement and evaluated in a noisier environment, which suggests that these strategies present a more stable training procedure for the proposed reinforcement learning framework.

\subsection{Dose-constrained methods}\label{sec:DoseConstraintExperiments}

In this study, we examine the two approaches to incorporate dose constraints introduced in Section \ref{sec:BudgetConstraint}. In the first case, the policy is a conditional Bernoulli distribution as in \cite{attia2022optimal}, where the design policy proposes patterns using a dose budget $B$ along a experiment, as described in \eqref{eq:budget_objective}. Training is carried out for budgets $B\in\{1,\ldots,8\}$, the number of experiment steps is $M=2$ and the image size is $N=4$. 

\begin{figure}[tbp!]
  \centering
  \includegraphics[width=0.8\textwidth]{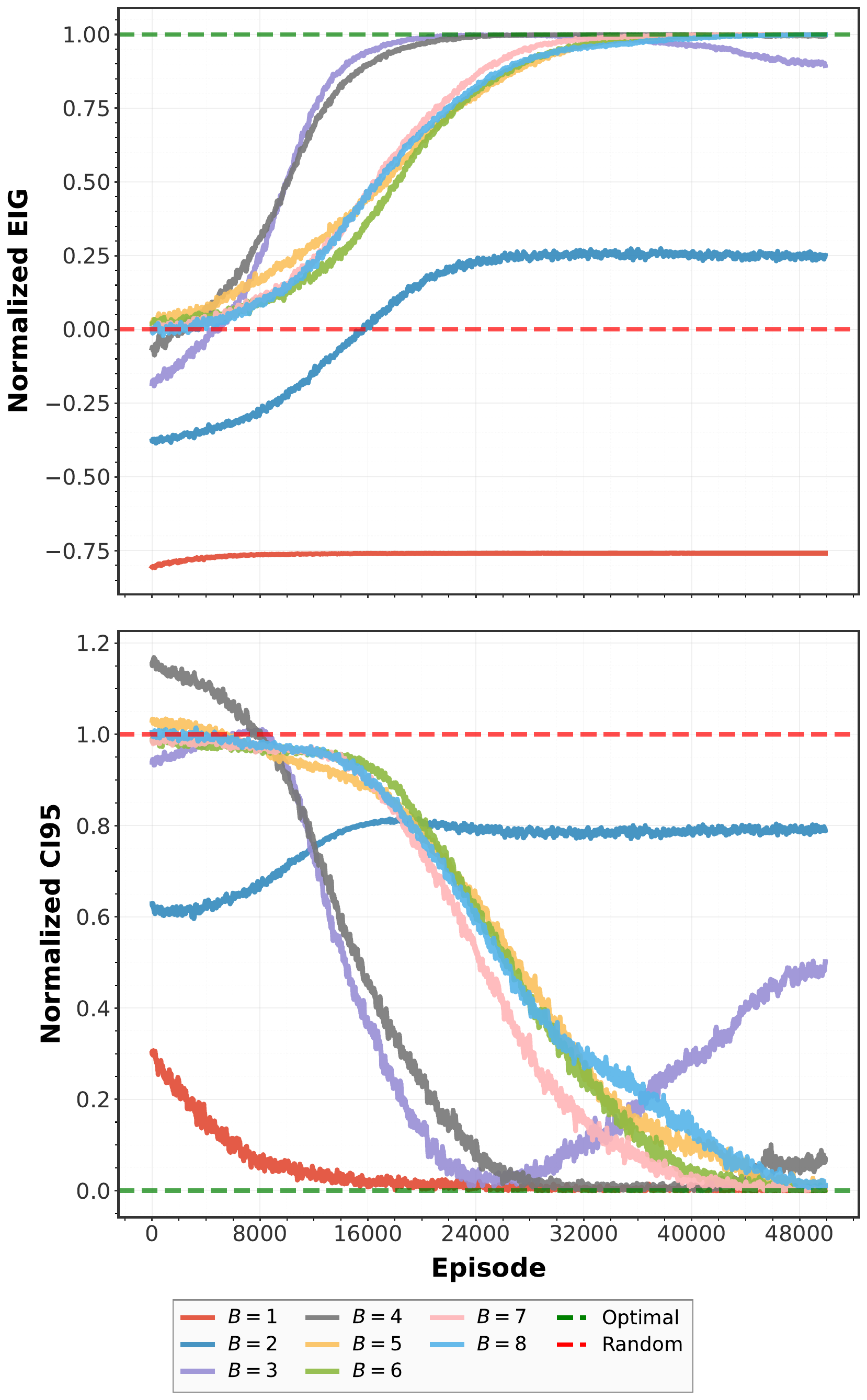}
  \caption{Total normalized EIG  under dose budget constraints $B \in \{1,\dots,8\}$ for conditional Bernoulli policies. First and second row display mean and $95\%$ confidence interval width, respectively. Random baseline, in red, is computed using a unconstrained Bernoulli policy. Optimal baseline is drawn in green.}
  \label{fig:Results_BudgetConstraint}
\end{figure}

Figure \ref{fig:Results_BudgetConstraint} shows the performance of constrained Bernoulli policies normalized by the optimal EIG and the strategy obtained using unconstrained Bernoulli policies without training, i.e., projecting patterns at random without dose constraints, as described in Section \ref{sec:ExperimentalSettings} (this results in an average dose of $4$ for this setup). We observe that only the trained policy with budget $B = 1$ cannot produce a policy that beats the unconstrained random policy, stabilizing at $\sim-0.75$ of the normalized EIG. $B = 2$ policies start below the unconstrained Bernoulli random baseline but after training, these policies stabilize around a normalized EIG of $\sim0.25$ with a large variability. This suggests that, even with a small budget, some of these policies produce dose-constrained strategies that are better than just drawing patterns from an unconstrained Bernoulli policy. Still, the budget is too small to reach optimal EIG strategies.

As we described in Section \ref{sec:OptimalEIGStrategies}, an adaptive policy, i.e., one that truly implements sOED, needs a budget of $B=3$ to reach optimal performance in this two-step experiment. We can see that the training curve indeed reaches an optimal EIG until episode 40,000 and then the performance drops in the last epochs. The policy presents the fastest learning of the optimal EIG strategy along with budget $B=4$, while this dose constraint is the first reaching almost zero variability around episode 25,000. Larger budget constraints $B\geq 5$ follow a similar trend, with a slower, more stable performance curve and low variability after the end of training.

% The case $B=3$ is the only one where achieving the EIG ceiling is possible only using adaptive strategies. It displays the fastest and most unstable learning dynamic of the budgets, when observing the confidence interval 95\%. For the $B\geq4$ case, fixed designs become available as a solution, resulting in the most stable and efficient balance of active and inactive pixels to discover a policy in terms of number of episodes.

\begin{figure}[tbp!]
    \centering
    \includegraphics[width=0.9\textwidth]{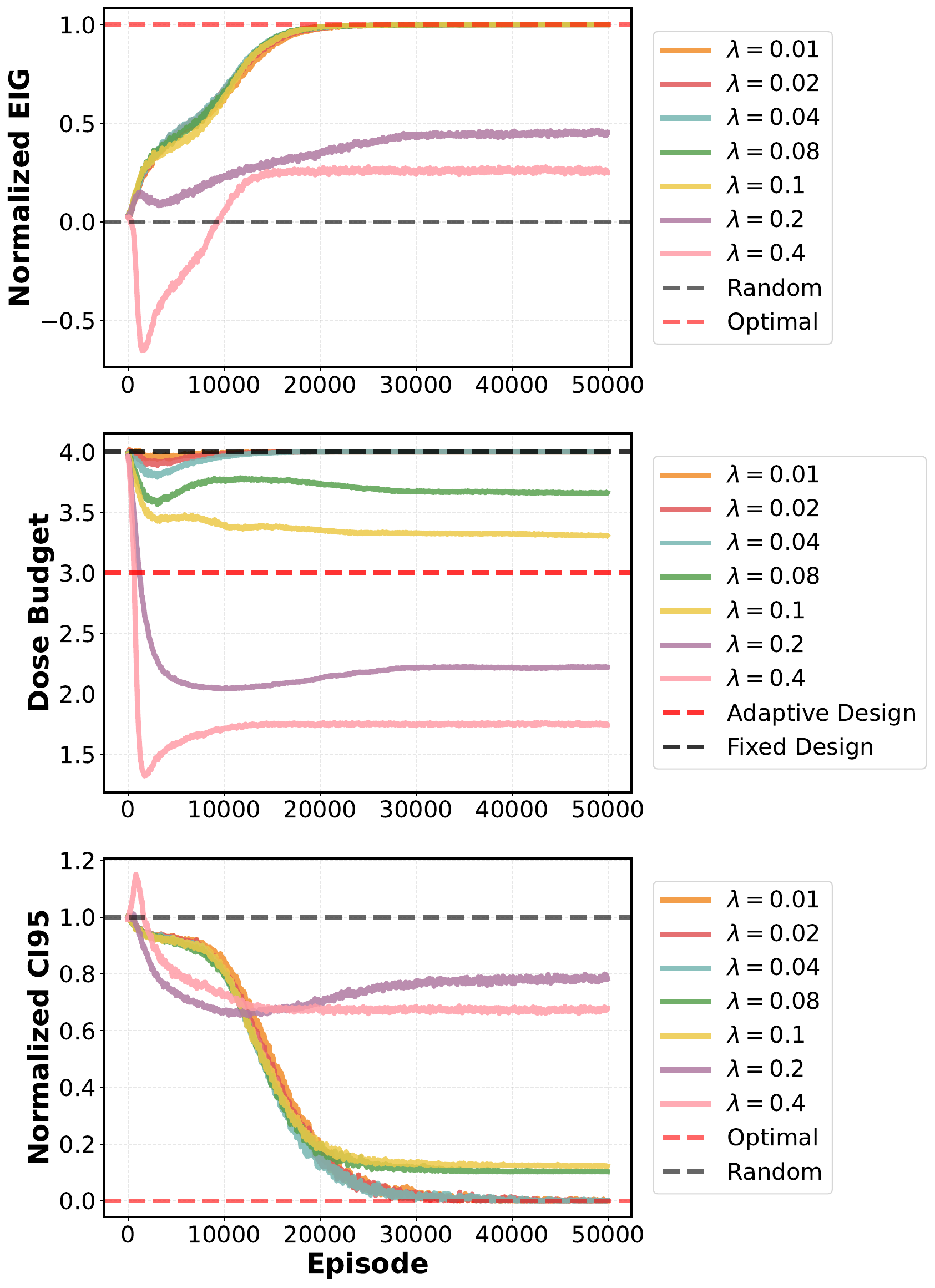} 
    \caption{From top to bottom, total normalized EIG, dose budgets $B$ and normalized $95\%$ confidence interval widths. Dose constraints $\lambda \in \{0.01, 0.02, 0.04, 0.08, 0.1, 0.2, 0.4\}$ are used, with random and optimal baseline strategies in gray and red, respectively. For the dose budget plot, baselines represent a fixed (gray) or an adaptive (red) policy design choice.}    
    \label{fig:lambda-sweep}
\end{figure}

Figure \ref{fig:lambda-sweep} illustrates how to incorporate the constraint as a penalty term, where $\lambda$ is the weight between the dose budget $B$ and the EIG to be optimized. We evaluate using $\lambda \in \{0.01, 0.02, 0.04, 0.08, 0.1, 0.2, 0.4\}$ and plot separately the EIG normalized between the optimal EIG strategy and the unconstrained random policy, the dose budget that each policy uses at the time of projecting patterns and the normalized 95\% confidence interval.

% The alternative to incorporate dose constraints is to maximize jointly the two-step EIG and the negative dose constraint to directly reward the objective in \eqref{eq:soft_constraint}. Using $\lambda \in \{0.01, 0.02, 0.04, 0.08, 0.1, 0.2, 0.4\}$ as dose budget constraint, Figure \ref{fig:lambda-sweep} illustrates separately the EIG and dose budgets mean values, as well as the 95\% confidence interval for normalized EIG. 

When using the dose as a soft constraint, different EIG performances are observed by the end of the learning process. For small $\lambda$ in the range of $[0.01, 0.04]$, the policies converge to an optimal EIG policy that proposes designs with $B = 4$. Intermediate $\lambda\in[0.8, 0.1]$ also display an optimal EIG behavior with a wider confidence interval and a mean budget $B\sim 3.5$. In the range of large $\lambda \in \{0.2, 0.4\}$, the dose constraint weights are so heavy that the resulting design no longer achieves optimal EIG. This constraint also proposes designs $B \sim 2$ that are suboptimal for the one-active pixel scenario and the confidence intervals are the widest. 

\section{Discussion}\label{sec:Discussion}

In Section \ref{sec:NumericalStudies} we illustrated the usage of the reinforcement learning framework to solve the Bayesian optimal experimental design of binary pattern for single-pixel imaging under a one-active pixel prior. For this methodology, we showed different aspects that affects the performance of the method and allow it to find optimal strategies incorporating constraints and measurement noise.

When choosing a particular parameterization for the policy, the learning curves in Section \ref{sec:PolicyArchitectureComparisonExperiments} display an inflection point that replicates over the experiments. We infer that, given the two steps we set in the proposed SPI experiment, namely projecting two binary patterns to the image of interest, learning the optimal design splits into two phases. As we examined intermediate pattern proposals along training, this behavior is likely caused by learning a bisection halving pattern in the first place and later on learning the remaining strategy. This phenomena can be related to the adaptive and fixed strategies introduced in Section \ref{sec:OptimalEIGStrategies}. At the first pattern, both strategies do not differ from each other, as they have a first bisection pattern with half of their pixels active. Moreover, the position of these active pixels is not crucial for the first pattern, but it is for the second one with respect to the first to achieve an optimal EIG step, since the input to the policy changed from being the prior in the first step to one of two possible posterior distributions in the second step. Then, the inflection point suggests a faster learning of the first pattern, where the policy training reaches a suboptimal EIG. Learning the second pattern takes more episodes for the policy due to the different possible posteriors as input for which a different optimal second pattern needs to be learned.

The sample-based state and reward studies in Section \ref{sec:SampleBasedState} show the potential of RL methods in large-scale imaging scenarios, where a Bayesian approach is only feasible via sampling methods. Here, we explored the effect of using state and reward estimators to learn optimal policies in the one-active pixel scenario. Our results show that training with a surprisingly small number of posterior samples is similarly fast compared to using the exact state or reward. The downside of this approach is the additional cost to simulating trajectories, which becomes more expensive in terms of compute due to the sampling times to compute the state or the reward estimators.
% Why are some sample based learning curves better than the "no-approximation" one? Maybe similar phenomena as in SGD: Adding the right amount of noise helps to overcome local minima in the loss landscape of the RL policy network

Lastly, training of policies under dose constraints in Section \ref{sec:DoseConstraintExperiments} demonstrate that better strategies than random Bernoulli patterns can be learned even for high dose constraints, i.e., low pattern activation budgets. These studies also directly demonstrate that our framework is able to learn truly adaptive policies, i.e., policies that solve the sOED problem \eqref{eq:sOED}, not just the a-priori OED problem \eqref{eq:OEDpri}: For both type of constraints, policies can only reach the optimal EIG with budget $B=3$ if they act adaptively to the first measurement, cf. Section \ref{sec:OptimalEIGStrategies}. %Studies for similar problems in sensor placement \cite{attia2024probabilistic} allow this sequential constraint in our experiments, although the scalability of this approach is computationally demanding.

\section{Conclusions}\label{sec:Conclusions}

This work presents a way to formulate the design proposal of single-pixel imaging binary patterns as a Bayesian sequential optimal experimental design problem. By using a reinforcement learning framework, we proposed different ways to parameterize a design policy, and illustrated how they can be used depending on the scale of the design size and the image size. For random binary patterns, if the image size is small, a categorical policy over all possible options is preferred over a Bernoulli one due to their faster training procedure.

To investigate the scaling from proof-of-concept to large scale computational imaging scenarios, we carried out sample-based approximations using numerical simulations. Surprisingly, we found that approximating the state and/or the reward with with sample-based estimators had only minor effects on the learning dynamics, even for very low numbers of samples.

We validated the robustness of the framework by learning designs when the measurements contain different noise levels. To accelerate training in this scenario, we showed that a curriculum learning strategy can be beneficial that uses a gradual noise scaling over the training episodes, simulating a degree of difficulty from simple to complex. 

Moreover, we studied two possible ways to impose constraints over the illumination dosage, enabling budget-constrained single-pixel experimental designs and proofing that the learned designs are truly adaptive. 

In this work, we deliberately limited our numerical studies to a simple, small-scale scenario, for which we could deduce optimal policies in an intuitive way. This allowed us to evaluate the learned policies in a thorough way and compute many quantities that are usually not tractable. In future work, we will extend this framework to other image priors and large-scale SPI scenarios.

\section*{Data availability}

The code supporting the findings of this study will be released publicly upon acceptance of the manuscript.

\section*{Ethics statement}

This study did not involve research with human participants, human data, human tissue, or live animals. Therefore, ethical approval was not required.

\section*{Funding}
The authors have no conflicts of interest to declare. This work was supported by the European Union (No. GA 101072354). Views and opinions expressed are, however, those of the author(s) only and do not necessarily reflect those of the European Union or the European Research Executive Agency. Neither the European Union nor the granting authority can be
held responsible for them.

\section*{Acknowledgements}

M.A.O. thanks L. Veenstra for insightful discussions on Bayesian inverse problems.
% This section is a list of funder names and grant numbers

% List author names and the contributions made to the article, using terms from the NISO Contributor Roles Taxonomy (CRediT) https://credit.niso.org

\bibliographystyle{IEEEtran}
\bibliography{bib}

\suppdata{Reward and state approximation}\label{sec:RewardStateApprox}

% When this is being done jointly, Figure \ref{fig:Results_StateApproximation_RewardApproximation} shows that when samples are reused, the combined degradation at low $N_s$ and low $N_r$ shows a noisier training performance than in the "No Approximation" case shown in Section \ref{sec:RewardApprox}, with their corresponding new optimal baseline. Higher $N_s$ converges to the no state approximation case curves recovering the curves obtained for the reward approximation, with an exponential requirement of $N_r$ samples to the exact scenario as observed in Figure \ref{fig:Results_StateExact_RewardApproximation}.

\begin{figure}[tbp!]
    \centering
    \includegraphics[width=0.9\linewidth]{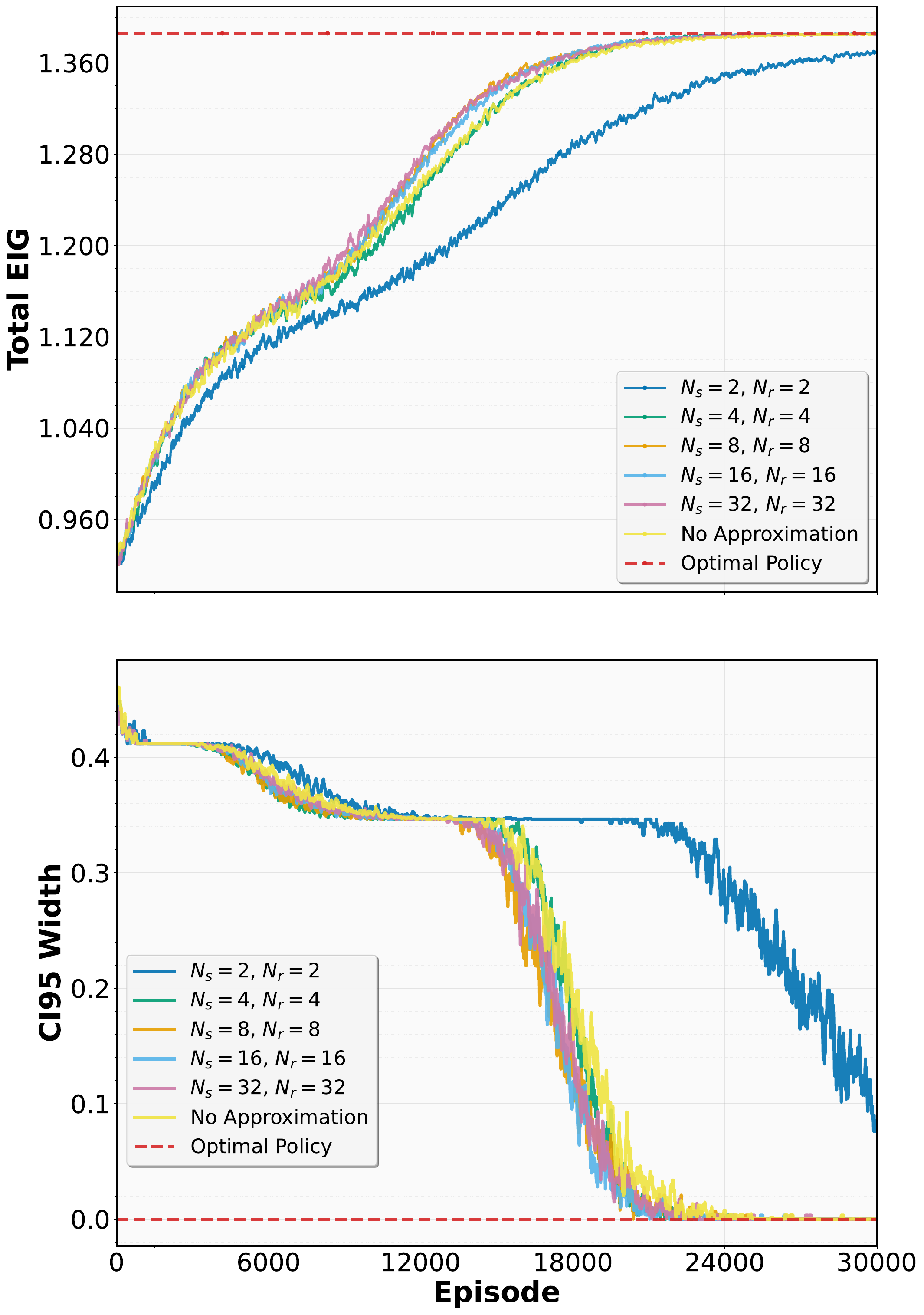}
    \caption{\textbf{Impact of reward approximation on policy learning.} Training curves showing normalized cumulative $EIG$ for categorical policy learning under a Monte Carlo reward approximation with $N_s, N_r \in \{2, 4, 8, 16, 32\}$ posterior samples. The same samples are used to compute both state and reward approximations. Curves are averaged across 100 independent policy training realizations.}
    \label{fig:Results_StateApproximation_RewardApproximation}
\end{figure}

Section \ref{sec:SampleBasedState} introduced separate studies varying the number of samples over the state and reward computation. In Figure \ref{fig:Results_StateApproximation_RewardApproximation} we study the effect of reusing the same samples for approximating the state and the reward. In comparison to the previous approximation studies, we observe a similar behavior to the reward approximation for $N_s,N_r \geq 4$, where the total EIG along training is similar with no different learning trend after the inflection point at episode 6,000. For $N_s,N_r = 2$, the learning curve is slower and the number of episodes is not enough for reaching the optimal EIG. With respect to the reward approximation alone, using $N_s,N_r = 2$ seems to have a stronger impact in learning the optimal strategy, suggesting that there is a strong interplay between state and reward approximation when samples are scarce.
 
\end{document}